\documentclass[sigconf, nonacm]{acmart}
\AtBeginDocument{%
  }
\usepackage{multirow}

\usepackage{CJKutf8}
\usepackage{xcolor}

\newcommand{\pmstd}[2]{%
    \ensuremath{%
        #1%
        \if\relax\detokenize{#2}\relax
        \else
            _{\scriptscriptstyle\pm #2}%
        \fi
    }%
}
\usepackage{xcolor}
\definecolor{bestorange}{RGB}{205,102,29}   

\newcommand{\second}[1]{\underline{#1}}
\usepackage{colortbl}

\begin{document}

\title{SOTER: A Generative Time-Series Foundation Model for Wearable Human Physiological Signals}

\author{Fangke Chen}
\affiliation{%
  \institution{School of Integrated Circuits, Zhejiang University}
  \city{Hangzhou}
  \country{China}
}
\affiliation{%
  \institution{Shanghai Innovation Institute}
  \city{Shanghai}
  \country{China}
}
\email{fkchen@zju.edu.cn}

\author{Sirry Chen}
\affiliation{%
  \institution{School of Data Science, Fudan University}
  \city{Shanghai}
  \country{China}
}
\affiliation{%
  \institution{Shanghai Innovation Institute}
  \city{Shanghai}
  \country{China}
}
\email{siyuanchen25@m.fudan.edu.cn}

\author{Wei Chen}
\affiliation{%
 \institution{School of Software Engineering,\\Huazhong University of Science and Technology}
 \city{Wuhan}
 \country{China}}
\email{lemuria\_chen@hust.edu.cn}

\author{Zhongyu Wei}
\authornote{Corresponding author.}
\affiliation{%
  \institution{School of Data Science, Fudan University}
  \city{Shanghai}
  \country{China}
}
\affiliation{%
  \institution{Shanghai Innovation Institute}
  \city{Shanghai}
  \country{China}
}
\email{zywei@fudan.edu.cn}



\begin{abstract}
Time-series foundation models have demonstrated strong cross-domain transfer, yet their common architectural assumptions remain poorly aligned with wearable physiological signals, which are multichannel, irregularly sampled, noisy, and governed by coupled continuous-time dynamics spanning distinct spectral scales. We present \textbf{SOTER}, a generative foundation model for wearable physiological time series that unifies cross-channel coupling, spectrum-guided expert specialization, and continuous-time latent evolution within a single pre-training framework. SOTER combines a spatial feature-aware backbone that models inter-signal dependencies, a power spectral density (PSD)-guided mixture-of-experts layer that routes representations to experts associated with fixed spectral bands through an inspectable, non-learned rule, and a neural controlled differential equation decoder that supports prediction and imputation at arbitrary timestamps. We pre-train SOTER on 226 billion time points from five public physiological datasets and evaluate the same pre-trained model across out-of-distribution zero-shot forecasting, frozen-encoder linear-probe classification, and continuous-time imputation on wearable benchmarks. SOTER achieves the best RMSE on 4 of 6 datasets and the best MAE on 5 of 6 in zero-shot forecasting, the highest average Macro-AUROC in classification, and the lowest imputation error on all six datasets at 75\% missingness. It further remains robust to additive acquisition noise, matching or surpassing baselines evaluated on clean inputs even under the strongest corruption. These results indicate that domain-specialized foundation models for wearable physiology benefit from jointly modeling channel structure, spectral scale, and continuous-time dynamics\footnote{Source code is available at \href{https://github.com/SII-fkchen/SOTER}{https://github.com/SII-fkchen/SOTER}}. 
\end{abstract}


%
\maketitle

\section{Introduction}
\label{sec:introduction}

Wearable sensors enable continuous and long-term monitoring of human
physiological signals. Such signals---including electroencephalography
(EEG) reflecting brain activity, photoplethysmography (PPG),
electrodermal activity (EDA), and blood volume pulse (BVP)---provide
complementary measurements of human physiological states~\cite{deng2025tardiff,harutyunyan2019multitask}. Capturing the underlying patterns within these signals through
large-scale representation learning can support downstream tasks such as
emotion recognition and seizure detection~\cite{nandini2025ensemble,ho2026artificial}, enabling earlier anomaly
warning and more timely clinical action~\cite{scheid2025development,kim2025seeing}. In practice, however,
physiological data collected across populations, devices, and
institutions often exhibit substantial distributional differences~\cite{wu2025term2note}, while privacy regulations such as GDPR further
constrain cross-region data aggregation~\cite{pulsar2023}. These
conditions motivate a domain-specialized time-series foundation model that can be
pre-trained on de-identified public data and transferred across datasets and tasks with limited downstream adaptation.

Time-series foundation models (TSFMs) provide a basis for such
unified modeling. Generalist TSFMs have advanced through multimodal inputs, flexible generative objectives, and sparse mixture-of-experts (MoE) architectures~\cite{wuaurora,sundial,timemoe}, but they are primarily designed around generic time-series assumptions. Most retain channel independent processing~\cite{Yuqietal-2023-PatchTST}, employ learned expert routing whose
specialization emerges implicitly during optimization, or operate on
discrete temporal grids. Recent medical and wearable TSFMs have begun to incorporate domain-specific designs: PaPaGei pre-trains unified representations for the PPG modality~\cite{pillai2025papagei}; MIRA incorporates neural ordinary differential equations (Neural ODEs)~\cite{neuralode} and continuous-time rotary position encoding for irregular observations~\cite{li2026mira}; and NormWear introduces channel-aware attention through a shared liaison token~\cite{normwear}. Nevertheless, existing approaches address only individual aspects of physiological modeling and do not jointly capture channel coupling, spectral specialization, and continuous-time generation within a unified pre-training framework. This gap reflects a broader mismatch between the assumptions of existing time-series models and the intrinsic structure of wearable physiological signals.

These physiological characteristics distinguish wearable signals from generic time series in three key ways. First, physiological channels are not independent signals; together, they provide complementary observations of a shared underlying state. State-dependent interactions exist among physiological systems such as the heart, respiratory system, and nervous system~\cite{bashan2012network,marzbanrad2020framework}. Processing channels independently may therefore discard coordinated patterns that cannot be recovered from any single signal alone. Second, physiological dynamics span heterogeneous temporal and spectral scales. These scales interact with physiological states: heart rate, blood pressure, and respiration exhibit cross-scale dependencies, while different physiological rhythms show characteristic spectral patterns under varying states~\cite{angelini2007multiscale,lin2020dynamic}. Uniform transformations across these regimes may obscure scale-specific structures and force shared parameters to model substantially different dynamics. Third, physiological processes evolve continuously, whereas wearable devices capture them at discrete, often irregular, and incomplete timestamps. Such acquisition patterns lead to missing observations and reduced signal reliability in real-world wearable monitoring~\cite{collins2021quantifying,hu2024lab}. Models restricted to fixed temporal grids cannot naturally query latent physiological states at arbitrary timestamps, limiting both future forecasting and causal imputation. Together, these properties motivate three requirements for a physiological foundation model: cross-channel coupling, spectrum-conditioned capacity allocation, and continuous-time state evolution.

To meet these requirements, we propose SOTER, a generative time-series
foundation model for wearable human physiological signals. SOTER employs
a spatial feature-aware backbone that preserves channel-specific temporal
modeling in its lower layers and restores cross-channel interaction over
high-level representations. It further introduces a power spectral
density (PSD)-guided mixture-of-experts layer, whose non-learned routing
rule associates experts with fixed spectral bands, providing
frequency-grounded and inspectable specialization without an auxiliary
routing loss. Finally, a terminal decoder based on neural controlled differential
equations (Neural CDEs) and Neural ODEs evolves the latent physiological
state to arbitrary query timestamps, thereby supporting both future
forecasting and causal imputation~\cite{neuralcde,neuralode}. These components are not independent
add-ons: they respectively model the structure, spectral scale, and temporal evolution of the same latent physiological process within a unified
generative objective. Our main contributions are as follows:

\begin{itemize}
    \item We introduce SOTER, a domain-specialized generative time-series
    foundation model for wearable physiological signals, pre-trained on
    226 billion time points from five public physiological datasets.
    SOTER enables unified transfer across zero-shot forecasting,
    frozen-representation classification, and continuous-time imputation
    from a single pre-trained backbone.

    \item We develop a physiology-aware architecture that integrates
    cross-channel coupling, PSD-guided expert specialization, and
    continuous-time latent evolution, aligning model capacity with the
    structural, spectral, and temporal properties of physiological
    signals.

    \item Extensive experiments across six out-of-distribution wearable
    datasets demonstrate that SOTER achieves consistent improvements in
    zero-shot forecasting, frozen representation evaluation, and
    continuous-time imputation.

    \item We further analyze the source of SOTER's transferability through
    corpus-matched pre-training comparisons and evaluations under noisy
    observations, limited supervision, and different pre-training
    exposures.
\end{itemize}





\section{Related Work}
\label{sec:related_work}

\subsection{Generalist Time-Series Foundation Models}
Generalist TSFMs are pre-trained on
large and heterogeneous corpora and have demonstrated strong zero-shot
transfer across forecasting domains
~\cite{gruver2023large,liu2024unitime,liu2024autotimes}. Recent
architectures span multimodal inputs that combine rendered time-series
images, text prompts, and raw signals~\cite{wuaurora}, flow-matching
objectives for flexible generative modeling~\cite{sundial}, group
attention for information sharing among related series
~\cite{ansari2025chronos}, and sparse mixture-of-experts architectures
for scalable capacity~\cite{timemoe,moiriamoe}. Other representative
families, including TimesFM~\cite{dastimesfm}, Moirai
~\cite{liu2025moirai}, MOMENT~\cite{MOMENT}, TiRex
~\cite{auer2026tirex}, TTM~\cite{ekambaram2024tiny}, and Lag-Llama
~\cite{lagllama}, further explore diverse forecasting and representation
learning designs. Although several of these models support multivariate
inputs or information sharing across related series, their dominant
assumptions remain generic: observations are represented on discrete
temporal grids, channel interactions are not designed around coupled
physiological systems, and expert specialization, when present, is
typically determined implicitly through learned routing. They therefore
provide strong general-purpose baselines, but do not directly address the
combined structural requirements of wearable physiological signals.

\subsection{Medical Time-Series Models} 
Deep learning has advanced medical time-series analysis across
forecasting, phenotyping, monitoring, diagnosis, and public-health
surveillance
~\cite{rubin2023forecasting,bridgeli,qin2023t,chen2022clustering,
zhang2023improving,li2022self,yang2021multi,ho2023self,
mciver2014wikipedia}. Most conventional approaches, however, are
optimized for a specific task, signal type, or clinical scenario,
limiting their reuse across heterogeneous physiological datasets and
downstream objectives. Recent medical and wearable TSFMs have begun to
improve cross-domain transfer. PaPaGei~\cite{pillai2025papagei} and
Pulse-PPG~\cite{pulseppg} learn representations specialized for PPG;
MIRA~\cite{li2026mira} incorporates Neural ODEs and
continuous-time rotary position encoding to accommodate irregular
observations; and NormWear~\cite{normwear} introduces a shared liaison
token to capture relationships among wearable sensor channels.

These models address important subsets of the physiological modeling
problem, but their capabilities remain complementary rather than
unified. Modality-specific models do not provide generative modeling
across heterogeneous physiological signals; continuous-time medical
models do not explicitly allocate capacity according to spectral
content; and channel-aware representation models are not designed to
support arbitrary-time forecasting and causal imputation through a
shared generative decoder. SOTER instead unifies high-level
cross-channel coupling, spectrum-conditioned expert specialization, and
continuous-time generative decoding within a single pre-trained time-series model
evaluated across forecasting, classification, and imputation.

\begin{figure*}[t!]
    \centering
    \includegraphics[width=\textwidth]{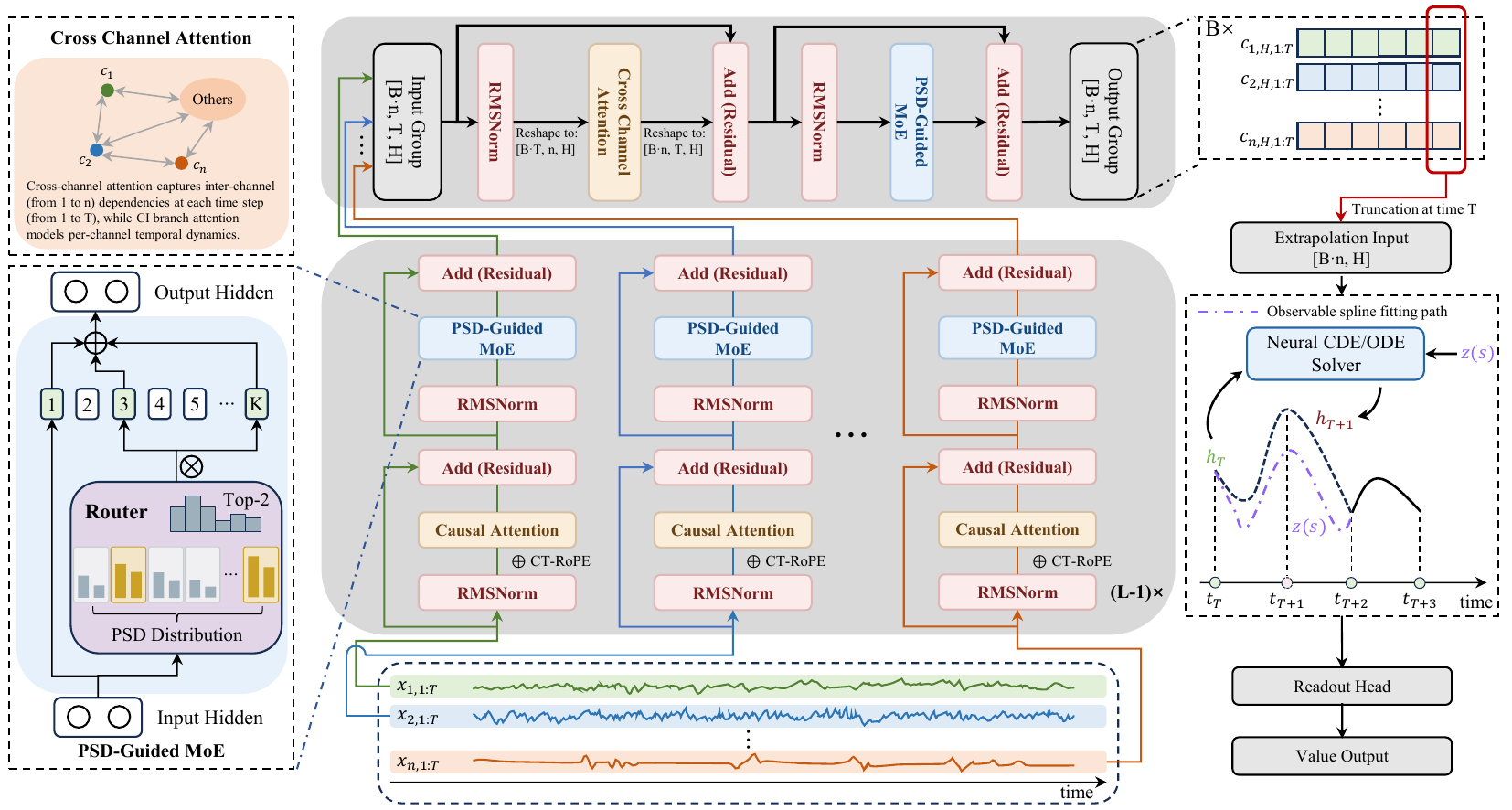}
    \caption{Overview of SOTER. The backbone consists of $L{-}1$ CI layers followed by one CD layer. The CI layers model the temporal dynamics of each channel independently, whereas the CD layer captures cross-channel dependencies over high-level representations. A terminal decoder based on Neural CDEs advances the latent state $h_T$ to an arbitrary query time using a control path $z(s)$ interpolated from the causal observation prefix. When the query lies beyond the observed interval and no control is available, the decoder reduces to a Neural ODE.}
    \Description{A left-to-right architecture diagram of SOTER. Multiple
    physiological channels are first processed independently by stacked
    causal temporal-attention layers with continuous-time positional
    encoding. Their representations are then reshaped for cross-channel
    attention. Each backbone block contains a power-spectral-density-guided
    mixture-of-experts module whose router selects experts associated with
    spectral bands. The final hidden state is passed to a continuous-time
    decoder, which operates as a controlled differential equation when an
    observed control path is available and as an ordinary differential
    equation for extrapolation beyond the observations.}
    \label{fig:overview}
\end{figure*}

\section{Methodology}
\label{sec:methodology}
\subsection{Problem Statement}
\label{sec:problem_statement}
We consider a collection of wearable physiological recordings, each
comprising $n$ channels observed at a shared, possibly irregularly spaced
sequence of timestamps $t_{1:T}=(t_1,\dots,t_T)$; the value of channel
$c\in\{1,\dots,n\}$ at timestamp $t_i$ is a scalar $x_{c,i}$. Given an observed history of wearable physiological recordings, SOTER
is pre-trained in a generative fashion to predict the value at the next
queried timestamp. An overview of the architecture is shown in
Figure~\ref{fig:overview}.

\subsection{SOTER Architecture}
\label{sec:soter_architecture}
\subsubsection{Continuous-Time Input Encoding}
Each scalar observation $x_{c,i}$ is first lifted to an $H$-dimensional
embedding through a gated projection:
\begin{equation}
h^{0}_{c,i}
=
\big(\mathrm{SiLU}(W_g x_{c,i})\big)
\odot
\big(W_e x_{c,i}\big),
\label{eq:input_embedding}
\end{equation}
where $W_g$ and $W_e$ map the single input channel to the hidden width,
and $\odot$ denotes the elementwise product. To avoid premature
cross-channel leakage, each channel is first treated as an independent
univariate sequence in the lower layers~\cite{Yuqietal-2023-PatchTST}. Because physiological signals are
frequently sampled at irregular intervals, discrete positional indices
poorly reflect their temporal geometry; SOTER therefore employs
continuous-time rotary positional encoding
(CT-RoPE)~\cite{li2026mira}, which rotates queries and keys by an angle
proportional to the absolute timestamp rather than the token position.

\subsubsection{Spatial Feature-Aware Backbone}

The backbone stacks $L{=}6$ pre-norm blocks, each employing
multi-head attention with a per-head dimension of $d$; the attention
operator is layer-dependent, with the first $L{-}1$ layers being
channel-independent (CI) and a single high-level layer being
channel-dependent (CD). Within each backbone block, the feed-forward
sublayer is implemented as the PSD-guided mixture-of-experts module, while the attention sublayer follows the CI/CD design described
above. In a CI layer, attention is applied independently to each channel
sequence, whose hidden states form
$U=[\,h_1;\dots;h_T\,]\in\mathbb{R}^{T\times H}$, with
$h_i\in\mathbb{R}^{H}$ denoting the representation at timestamp $t_i$,
using causal self-attention modulated by CT-RoPE:
\begin{equation}
\begin{gathered}
\mathcal{A}^{\mathrm{CI}}_{ij}
=
h_i^{\top}
W_q
\mathbf{R}_{\Theta,\,t_i-t_j}
W_k^{\top}
h_j,
\\[2pt]
\mathrm{Attention}^{\mathrm{CI}}(\mathbf{U})
=
\mathrm{Softmax}\!\left(
\frac{\mathrm{Mask}(\mathcal{A}^{\mathrm{CI}})}
{\sqrt{d}}
\right)
\mathbf{U}W_v.
\end{gathered}
\label{attentionci}
\end{equation}
Here, $W_q,W_k,W_v\in\mathbb{R}^{H\times d}$ are the query, key, and
value projections,
$\mathbf{R}_{\Theta,\,t}\in\mathbb{R}^{d\times d}$ is the CT-RoPE
rotary matrix whose rotation angle is determined by the absolute
timestamp $t$, and $\mathrm{Mask}(\cdot)$ is the causal mask that forbids
attending to future timestamps. Equations~\eqref{attentionci}
and~\eqref{attentioncd} are written for a single head; the standard
$M$-head extension ($Md=H$, with outputs mixed by
$W_O\in\mathbb{R}^{Md\times H}$) is omitted.

In the CD layer, cross-channel coupling is restored, which is essential
because the human body behaves as a strongly coupled dynamical
system~\cite{bashan2012network}. After channel-independent temporal modeling, representations are
regrouped at each timestamp to restore cross-channel interaction.
For a given timestamp, the channel representations are denoted as
$\tilde{U}=[\tilde h_1;\dots;\tilde h_n]\in\mathbb{R}^{n\times H}$,
where $\tilde h_a$ represents the $a$-th channel. Full non-causal attention is then computed over the
$n$ channel tokens at each timestamp:
\begin{equation}
\begin{gathered}
\mathcal{A}^{\mathrm{CD}}_{ab}
=
\tilde{h}_a^{\top}
W_{\tilde q}
W_{\tilde k}^{\top}
\tilde{h}_b,
\\[2pt]
\mathrm{Attention}^{\mathrm{CD}}(\tilde{\mathbf{U}})
=
\mathrm{Softmax}\!\left(
\frac{\mathcal{A}^{\mathrm{CD}}}{\sqrt{d}}
\right)
\tilde{\mathbf{U}}W_{\tilde v}.
\end{gathered}
\label{attentioncd}
\end{equation}
Here, $a,b\in\{1,\dots,n\}$ index the channels, and
$W_{\tilde q},W_{\tilde k},W_{\tilde v}\in\mathbb{R}^{H\times d}$ are
the query, key, and value projections of the CD layer. No positional
encoding or causal mask is applied along the channel axis, since
channels carry no intrinsic ordering. Confining cross-channel attention
to a single top layer keeps the $\mathcal{O}(n^2)$ cost over channels to
one block, while letting it operate over refined, rather than shallow,
temporal representations. When $n{=}1$, this layer falls back to
standard channel-independent temporal attention, so SOTER transfers
across recordings with different channel counts.

\subsubsection{PSD-Guided Mixture-of-Experts}
Prevailing MoE-based foundation models route tokens
with a learned gate regularized by an auxiliary load-balancing
loss~\cite{timemoe,li2026mira}, so that expert assignment follows no
explicit criterion and emerges only from training. SOTER instead derives routing directly from the spectral content of the
latent representation, using a PSD-inspired statistic to provide an
explicit frequency-aware criterion for expert assignment.For each latent trajectory, we maintain a strictly causal, $N$-point
prefix discrete Fourier transform (DFT)~\cite{fftori} over the latent temporal index.
Here, $t$ denotes the current latent position and $\tau$ denotes the
historical index within the causal prefix. We then average its magnitude
over the hidden dimension:
\begin{equation}
\begin{aligned}
X_t^{(d)}(k)
&=
\sum_{\tau\leq t}
h_{\tau}^{(d)}
e^{-\mathrm{j}\,2\pi k\tau/N},p_t(k)=
\frac{1}{H}
\sum_{d=1}^{H}
\left|X_t^{(d)}(k)\right|.
\end{aligned}
\label{eq:prefix_dft}
\end{equation}

The transform is taken over latent temporal indices, i.e., on the normalized
frequency axis of the latent trajectory. Absolute and possibly
irregular timing is handled separately by CT-RoPE and the CDE decoder, so
the router needs only a sampling-rate-invariant spectral-shape statistic
rather than a physical-frequency spectrum, which keeps expert assignment
consistent across the heterogeneous pre-training corpora.

The $\lfloor N/2\rfloor{+}1$ frequency bins are partitioned into $E$
contiguous bands $\{\mathcal{B}_e\}$. The routing logit of expert $e$ is the aggregated spectral magnitude
within $\mathcal{B}_e$, from
which the top-$2$ experts are selected and weighted:
\begin{equation}
\begin{aligned}
g_{t,e}
&=
\sum_{k\in\mathcal{B}_e} p_t(k),
\\[4pt]
r_{t,e}
&=
\frac{\exp(g_{t,e})}
{\sum_{e'=1}^{E}\exp(g_{t,e'})}
\,
\mathbf{1}
\left[
e\in\mathrm{TopK}(g_t,2)
\right].
\end{aligned}
\label{eq:psd_routing}
\end{equation}
The band-wise statistic $g_{t,e}$ aggregates spectral magnitudes within
$\mathcal{B}_e$; magnitude accumulation avoids excessive amplification
of dynamic ranges across bands. Because
the routing statistic is computed without gradients and the prefix
transform depends only on past tokens, the assignment is deterministic,
strictly causal, and free of any auxiliary balancing loss, so each expert
is tied to a fixed spectral band by construction.

A shared expert acting on the sequence-averaged representation is finally
added as a global residual:
\begin{equation}
\mathrm{MoE}(h_t)
=
\mathcal{E}_{\mathrm{sh}}
\left(
\frac{1}{T}
\sum_{\tau}h_{\tau}
\right)
+
\sum_{e\in\mathrm{TopK}(g_t,2)}
r_{t,e}\,
\mathcal{E}_e(h_t).
\label{eq:moe_output}
\end{equation}

This shared branch is a sequence-level residual rather than a per-token
causal path: it adds the mean over the window's $T$ tokens to every
position. Since the prediction target $t_{T+1}$ lies strictly after the
entire window, every token entering this aggregate belongs to the causal
prefix of the target, so the shared branch introduces no future leakage
with respect to the prediction and remains consistent with the strictly
causal prefix-DFT routing and causal attention used throughout.

\subsubsection{Neural CDE Extrapolation Decoder}
Autoregressive architectures generate under causal masking and cannot
access the target timestamp at inference. To extrapolate to arbitrary
timestamps, SOTER takes the state $h_T$ at the last observed time $t_T$
of each channel and evolves it to the target time $t_{T+1}$ through a
controlled continuous-time field. Unlike a Neural
ODE~\cite{neuralode}, which depends only on time and state, the vector
field of SOTER additionally receives an external control path $z(s)$:
\begin{equation}
\frac{\mathrm{d}h(s)}{\mathrm{d}s}
=
f_{\theta}\big(s,h(s),z(s)\big),
h(t_T)=h_T,
s\in[t_T,t_{T+1}].
\label{eq:cde_dynamics}
\end{equation}
Here, $f_{\theta}$ is a time-augmented multilayer perceptron (MLP) taking
the concatenation $[\,s;h(s);z(s)\,]$ as input. The target state follows
from numerical integration with an adaptive Dormand--Prince 5(4) solver (dopri5):
\begin{equation}
h(t_{T+1})
=
h_T
+
\int_{t_T}^{t_{T+1}}
f_{\theta}\big(s,h(s),z(s)\big)
\,\mathrm{d}s,
\label{eq:integration}
\end{equation}
from which a linear readout head produces the prediction
$\hat{x}_{T+1}=W_o h(t_{T+1})$.

During inference, the same continuous-time decoder supports two
querying regimes through the control path $z(s)$. When the target timestamp lies within the span of
observations, the decoder operates as a Neural
CDE-style controlled regime~\cite{neuralcde}: the observed points are interpolated into $z(s)$ by natural cubic splines, and the latent dynamics are steered by the
observable signal. Although a natural cubic spline is a global
interpolant, the spline for each queried point is fitted only on
observations strictly before the prediction origin---later samples are
physically excluded from the input rather than down-weighted by
masking---so the control path remains strictly causal, consistent with
the online streaming imputation setting. When no control is available beyond the observation span, we set \(z(s)\equiv 0\), and the decoder reduces exactly to a time-augmented Neural ODE~\cite{neuralode}. We provide more details in Appendix~\ref{app:cde_ode}. A single continuous-time decoder thus
unifies strictly causal extrapolation and observation-guided
interpolation, which underpins the transferability of a single frozen
SOTER across both forecasting and imputation.

\begin{table*}[t!]
    \caption{Zero-shot forecasting on six out-of-distribution wearable
    datasets. Values are reported as
    $\mathrm{mean}_{\pm\mathrm{std}}$ over four input/output window pairs
    (48/24,72/36,96/48,128/64); Timer is evaluated on the latter
    two only, due to its native configuration. Entries are rescaled by
    $10^{2}$. Paras denotes parameters activated per inference step
    (millions), and SC and MC denote single- and multi-channel datasets.
    The best result is indicated by bold style with an asterisk ($^{*}$), while the second-best result is underlined.}
    \label{tab:main_results}
    \centering
    \scriptsize
    \renewcommand{\arraystretch}{1.1}
    \setlength{\tabcolsep}{0.58mm}
    \resizebox{\textwidth}{!}{%
    \begin{tabular}{l c c *{12}{c}}
        \toprule
        \multirow{2}{*}{Method}
        & \multirow{2}{*}{Paras}
        & \multirow{2}{*}{Source}
        & \multicolumn{2}{c}{\shortstack{HeartRate\\(SC)}}
        & \multicolumn{2}{c}{\shortstack{MIT-BIH\\(MC)}}
        & \multicolumn{2}{c}{\shortstack{ScientiSST\\MOVE (MC)}}
        & \multicolumn{2}{c}{\shortstack{IEEE PPG\\(MC)}}
        & \multicolumn{2}{c}{\shortstack{MMASH\\(MC)}}
        & \multicolumn{2}{c}{\shortstack{WDD\\(MC)}} \\
        \cmidrule(lr){4-5}
        \cmidrule(lr){6-7}
        \cmidrule(lr){8-9}
        \cmidrule(lr){10-11}
        \cmidrule(lr){12-13}
        \cmidrule(lr){14-15}
        & & &
        RMSE & MAE &
        RMSE & MAE &
        RMSE & MAE &
        RMSE & MAE &
        RMSE & MAE &
        RMSE & MAE \\
        \midrule
        Timer           & 84.14M & ICML'24 & \pmstd{8.90}{1.26} & \pmstd{4.85}{0.52} & \pmstd{6.11}{0.25} & \pmstd{3.64}{0.01} & \pmstd{15.60}{0.01} & \pmstd{11.92}{0.01} & \pmstd{12.92}{0.54} & \pmstd{8.71}{0.39} & \pmstd{2.91}{0.28} & \pmstd{2.19}{0.13} & \pmstd{3.32}{0.10} & \pmstd{2.70}{0.09} \\
        TTM             & 0.805M & NIPS'24 & \pmstd{5.05}{0.84} & \pmstd{2.82}{0.52} & \pmstd{4.14}{0.11} & \pmstd{2.06}{0.10} & \pmstd{9.60}{0.09} & \pmstd{7.43}{0.21} & \pmstd{9.15}{0.20} & \pmstd{6.06}{0.24} & \pmstd{1.48}{0.44} & \pmstd{0.94}{0.22} & \pmstd{2.07}{0.24} & \pmstd{1.60}{0.25} \\
        Lag-Llama       & 2.451M & ARXIV'23 & \pmstd{19.96}{0.32} & \pmstd{15.63}{0.13} & \pmstd{48.93}{0.09} & \pmstd{48.64}{0.09} & \pmstd{51.02}{0.37} & \pmstd{46.92}{0.40} & \pmstd{51.75}{0.09} & \pmstd{50.69}{0.10} & \pmstd{18.37}{0.09} & \pmstd{17.59}{0.07} & \pmstd{43.75}{0.18} & \pmstd{41.45}{0.01} \\
        TiRex           & 35.29M & NIPS'25 & \pmstd{5.22}{1.25} & \pmstd{2.49}{0.55} & \pmstd{4.15}{0.32} & \pmstd{2.02}{0.12} & \pmstd{5.82}{0.52} & \pmstd{4.05}{0.51} & \pmstd{8.93}{0.36} & \pmstd{5.35}{0.27} & \pmstd{0.37}{0.25} & \pmstd{0.16}{0.13} & \pmstd{\second{0.98}}{0.07} & \pmstd{0.77}{0.09} \\
        MOMENT-Large    & 341.3M & \multirow{3}{*}{ICML'24}  & \pmstd{6.33}{0.40} & \pmstd{3.36}{0.23} & \pmstd{4.23}{0.11} & \pmstd{2.21}{0.05} & \pmstd{9.74}{1.17} & \pmstd{7.90}{0.92} & \pmstd{10.60}{0.72} & \pmstd{6.61}{0.50} & \pmstd{0.69}{0.31} & \pmstd{0.31}{0.14} & \pmstd{1.28}{0.24} & \pmstd{0.70}{0.16} \\
        MOMENT-Base     & 109.6M &                     & \pmstd{6.94}{0.31} & \pmstd{3.82}{0.34} & \pmstd{4.80}{0.30} & \pmstd{2.52}{0.07} & \pmstd{9.85}{1.16} & \pmstd{7.76}{0.96} & \pmstd{10.91}{0.82} & \pmstd{6.76}{0.55} & \pmstd{0.65}{0.30} & \pmstd{0.30}{0.13} & \pmstd{1.33}{0.28} & \pmstd{0.74}{0.16} \\
        MOMENT-Small    & 35.34M &                     & \pmstd{6.59}{0.34} & \pmstd{3.47}{0.34} & \pmstd{4.38}{0.10} & \pmstd{2.30}{0.05} & \pmstd{9.47}{1.22} & \pmstd{7.55}{0.98} & \pmstd{11.06}{0.75} & \pmstd{6.84}{0.51} & \pmstd{0.71}{0.30} & \pmstd{0.32}{0.15} & \pmstd{1.28}{0.27} & \pmstd{0.69}{0.15} \\
        MoiraiMoE-Base  & 935.1M & \multirow{2}{*}{ICML'25}  & \pmstd{7.00}{0.81} & \pmstd{3.43}{0.71} & \pmstd{5.47}{0.55} & \pmstd{2.50}{0.23} & \pmstd{13.18}{0.71} & \pmstd{9.25}{0.58} & \pmstd{11.68}{0.74} & \pmstd{7.06}{0.20} & \pmstd{0.79}{0.52} & \pmstd{0.30}{0.19} & \pmstd{1.27}{0.10} & \pmstd{1.05}{0.04} \\
        MoiraiMoE-Small & 117.0M &                     & \pmstd{7.30}{0.80} & \pmstd{3.65}{0.67} & \pmstd{5.25}{0.38} & \pmstd{2.41}{0.16} & \pmstd{13.43}{0.68} & \pmstd{9.60}{0.30} & \pmstd{12.02}{0.70} & \pmstd{7.16}{0.09} & \pmstd{0.77}{0.44} & \pmstd{0.30}{0.16} & \pmstd{1.25}{0.12} & \pmstd{1.04}{0.07} \\
        Moirai2.0-Small & 11.39M & Salesforce'26 & \pmstd{5.33}{0.28} & \pmstd{2.65}{0.40} & \pmstd{4.59}{0.47} & \pmstd{2.16}{0.17} & \pmstd{8.12}{0.44} & \pmstd{4.23}{0.13} & \pmstd{9.19}{0.81} & \pmstd{5.61}{0.38} & \pmstd{0.32}{0.27} & \pmstd{0.12}{0.10} & \pmstd{1.07}{0.13} & \pmstd{0.89}{0.06} \\
        TimesFM 1.0     & 203.6M & ICML'24 & \pmstd{6.04}{1.00} & \pmstd{2.78}{0.37} & \pmstd{4.38}{0.09} & \pmstd{2.14}{0.06} & \pmstd{10.74}{1.05} & \pmstd{7.94}{0.61} & \pmstd{8.60}{0.62} & \pmstd{5.30}{0.09} & \pmstd{0.33}{0.05} & \pmstd{0.11}{0.03} & \pmstd{1.03}{0.14} & \pmstd{0.72}{0.08} \\
        TimesFM 2.0     & 498.8M & \multirow{2}{*}{Google'25}  & \pmstd{6.46}{0.79} & \pmstd{3.07}{0.37} & \pmstd{4.42}{0.37} & \pmstd{2.09}{0.11} & \pmstd{11.02}{0.76} & \pmstd{8.09}{0.34} & \pmstd{9.24}{0.88} & \pmstd{5.67}{0.13} & \pmstd{\second{0.23}}{0.09} & \pmstd{\second{0.08}}{0.04} & \pmstd{1.00}{0.12} & \pmstd{0.69}{0.03} \\
        TimesFM 2.5     & 231.3M &                     & \pmstd{5.55}{0.54} & \pmstd{2.48}{0.27} & \pmstd{4.11}{0.32} & \pmstd{1.97}{0.11} & \pmstd{10.87}{1.13} & \pmstd{7.86}{0.76} & \pmstd{8.79}{0.77} & \pmstd{5.35}{0.18} & \pmstd{\mathbf{0.08^{*}}}{0.04} & \pmstd{\mathbf{0.03^{*}}}{0.02} & \pmstd{1.05}{0.16} & \pmstd{0.81}{0.10} \\
        TimeMoE-Large   & 453.2M & \multirow{2}{*}{ICLR'25}  & \pmstd{\second{1.68}}{0.67} & \pmstd{\second{1.16}}{0.32} & \pmstd{\second{1.82}}{0.11} & \pmstd{\second{1.00}}{0.03} & \pmstd{\second{3.10}}{0.09} & \pmstd{\second{1.95}}{0.07} & \pmstd{\mathbf{1.77^{*}}}{0.07} & \pmstd{\second{1.08}}{0.01} & \pmstd{3.06}{0.24} & \pmstd{2.61}{0.20} & \pmstd{1.57}{0.07} & \pmstd{1.21}{0.03} \\
        TimeMoE-Base    & 113.3M &                     & \pmstd{1.85}{0.33} & \pmstd{1.30}{0.22} & \pmstd{2.15}{0.17} & \pmstd{1.39}{0.21} & \pmstd{3.51}{0.13} & \pmstd{2.22}{0.06} & \pmstd{\second{1.79}}{0.07} & \pmstd{1.14}{0.02} & \pmstd{3.31}{0.23} & \pmstd{2.65}{0.16} & \pmstd{2.16}{0.13} & \pmstd{1.67}{0.07} \\
        Chronos2-Base   & 119.5M & \multirow{2}{*}{Amazon'25}  & \pmstd{5.65}{1.01} & \pmstd{2.49}{0.40} & \pmstd{4.80}{1.19} & \pmstd{2.11}{0.34} & \pmstd{7.21}{0.70} & \pmstd{3.88}{0.25} & \pmstd{9.57}{1.16} & \pmstd{5.79}{0.54} & \pmstd{\second{0.23}}{0.16} & \pmstd{0.10}{0.05} & \pmstd{1.09}{0.07} & \pmstd{0.63}{0.15} \\
        Chronos2-Small  & 27.93M &  & \pmstd{5.67}{0.30} & \pmstd{2.73}{0.32} & \pmstd{4.54}{0.80} & \pmstd{2.11}{0.34} & \pmstd{10.45}{1.52} & \pmstd{7.17}{1.13} & \pmstd{9.47}{1.05} & \pmstd{5.77}{0.56} & \pmstd{0.29}{0.17} & \pmstd{0.11}{0.06} & \pmstd{1.10}{0.10} & \pmstd{0.64}{0.14} \\
        Sundial-Base    & 128.3M & ICML'25 & \pmstd{5.75}{1.01} & \pmstd{2.77}{0.33} & \pmstd{4.12}{0.19} & \pmstd{2.00}{0.05} & \pmstd{10.75}{0.58} & \pmstd{7.47}{1.49} & \pmstd{8.40}{0.48} & \pmstd{5.22}{0.07} & \pmstd{1.14}{0.32} & \pmstd{0.97}{0.25} & \pmstd{1.38}{0.14} & \pmstd{0.95}{0.14} \\
        Aurora          & 210.8M & ICLR'26 & \pmstd{7.13}{1.04} & \pmstd{3.97}{0.43} & \pmstd{5.10}{0.22} & \pmstd{2.61}{0.09} & \pmstd{13.79}{1.75} & \pmstd{10.69}{1.47} & \pmstd{10.30}{0.45} & \pmstd{6.52}{0.20} & \pmstd{1.99}{0.27} & \pmstd{0.81}{0.07} & \pmstd{1.17}{0.17} & \pmstd{\second{0.62}}{0.05} \\
        MIRA-Large      & 339.9M & NIPS'25 & \pmstd{2.66}{1.56} & \pmstd{1.85}{1.11} & \pmstd{2.20}{0.14} & \pmstd{1.54}{0.12} & \pmstd{4.76}{0.16} & \pmstd{3.73}{0.12} & \pmstd{2.57}{0.06} & \pmstd{2.04}{0.01} & \pmstd{6.23}{0.24} & \pmstd{5.49}{0.27} & \pmstd{3.36}{0.03} & \pmstd{3.00}{0.03} \\
        \midrule
        SOTER           & 20.29M & / & \pmstd{\mathbf{1.41^{*}}}{0.15} & \pmstd{\mathbf{0.95^{*}}}{0.11} & \pmstd{\mathbf{1.29^{*}}}{0.22} & \pmstd{\mathbf{0.62^{*}}}{0.10} & \pmstd{\mathbf{2.26^{*}}}{0.12} & \pmstd{\mathbf{1.17^{*}}}{0.04} & \pmstd{2.19}{0.19} & \pmstd{\mathbf{1.05^{*}}}{0.05} & \pmstd{1.26}{0.26} & \pmstd{0.89}{0.15} & \pmstd{\mathbf{0.73^{*}}}{0.10} & \pmstd{\mathbf{0.60^{*}}}{0.09} \\
        \bottomrule
    \end{tabular}%
    }
\end{table*}

\subsection{Model Training}
\label{sec:model_training}

SOTER is pre-trained on the next-timestamp prediction task over a
wearable physiological corpus of approximately 226 billion time points,
assembled from five publicly available and de-identified datasets:
MIMIC-III Waveform~\cite{mimic3wdb,johnson2016mimic},
Sleep-EDF~\cite{sleepedf},
PTB-XL~\cite{wagner2020ptb,PhysioNet-ptb-xl-1.0.3},
WESAD~\cite{wesad}, and
Chapman-ECG~\cite{chapmanecg,PhysioNet-chapmanecg-arrhythmia-1.0.0}.
The training objective is a masked Huber loss~\cite{huber1992robust},
chosen for its robustness to the motion and contact artifacts common in
wearable acquisition. Training alternates per step, with equal probability, between two
decoding regimes that share the same next-timestamp target. In the
full-history regime, the decoder extrapolates from fully observed inputs
as a pure Neural ODE~\cite{neuralode}; in the missingness regime, we
randomly mask $15\%$ of historical observations while preserving their
timestamps. The masked values are zero-filled for backbone processing, while the remaining observations construct the spline control path, enabling a Neural CDE-style continuous-time decoder driven by observed signals~\cite{neuralcde}. Alternating between them
jointly activates both decoder modes and additionally provides
robustness pre-training under corrupted observations. We train SOTER for one epoch with a maximum sequence length of $512$ on
$8\times$ NVIDIA H200 GPUs, taking approximately $312$ hours. The model
comprises $49.79$M parameters, of which $20.29$M are activated per
inference step under top-$2$ routing; key hyperparameters are provided
in the Appendix~\ref{app:pretraining}.

\section{Experiments and Analysis}
\subsection{Experimental Setup}
\label{sec:experimental_setup}
We evaluate SOTER on six publicly available and de-identified wearable
physiological datasets: HeartRate~\cite{heartrate} (single-channel),
MIT-BIH~\cite{mitbih}, ScientiSST MOVE~\cite{scientisst-move1},
IEEE PPG~\cite{ieeeppg}, MMASH~\cite{PhysioNet-mmash,mmash}, and
WDD~\cite{PhysioNet-wdd,wdd} (multi-channel). All datasets use a
leave-one-subject-out (LOSO) split. Per-channel MinMax normalization is
estimated exclusively from the training split, preventing distributional
information from leaking from the test samples. We organize the compared methods into three groups. General-purpose
TSFMs include Aurora~\cite{wuaurora}, Sundial~\cite{sundial},
Chronos-2~\cite{ansari2025chronos}, TimeMoE~\cite{timemoe},
TimesFM~\cite{dastimesfm}, Moirai 2.0~\cite{liu2025moirai},
Moirai-MoE~\cite{moiriamoe}, MOMENT~\cite{MOMENT},
TiRex~\cite{auer2026tirex}, TTM~\cite{ekambaram2024tiny},
Timer~\cite{timer}, and Lag-Llama~\cite{lagllama}. Medical and wearable
TSFMs include MIRA~\cite{li2026mira} and NormWear~\cite{normwear},
while TFC~\cite{tfc} serves as a physiological representation-learning
baseline. For each downstream task, we select the subset of methods
compatible with its evaluation protocol: discriminative representation
models for classification, and models with generative or reconstruction
capabilities for forecasting and imputation, as detailed in the
corresponding result tables. 

All methods use their officially released checkpoints without updating
the pre-trained backbones and share the same input windows, normalization
procedure, and test splits as SOTER. Timer is evaluated on the two
longest windows only because of its native configuration. All experiments
are conducted on a single NVIDIA RTX 4090 GPU.

\begin{table*}[t!]
    \caption{Performance on the downstream classification task. We report
    Macro-F1 (\%) and Macro-AUROC on three wearable datasets, together
    with the average Macro-AUROC across datasets. Mean and standard
    deviation over three runs. The best scores are highlighted by bold style, and the second-best scores are underlined.}
    \label{tab:classification}
    \centering
    \small
    \renewcommand{\arraystretch}{1.0}
    \setlength{\tabcolsep}{1.2mm}
    \begin{tabular}{l *{7}{c} c}
        \toprule
        \multirow{2}{*}{Methods}
        & \multicolumn{2}{c}{WDD (3-Classes)}
        & \multicolumn{2}{c}{ScientiSST MOVE (7-Classes)}
        & \multicolumn{2}{c}{MMASH (6-Classes)}
        & \multirow{2}{*}{\shortstack{Average of\\Macro-AUROC}} \\
        \cmidrule(lr){2-3}
        \cmidrule(lr){4-5}
        \cmidrule(lr){6-7}
        & Macro-F1
        & Macro-AUROC
        & Macro-F1
        & Macro-AUROC
        & Macro-F1
        & Macro-AUROC
        & \\
        \midrule
        Random Baseline & \pmstd{32.90}{0.42} & \pmstd{50.00}{0.00} & \pmstd{7.40}{0.49} & \pmstd{50.00}{0.00} & \pmstd{13.61}{0.16} & \pmstd{50.00}{0.00} & 50.00 \\
        \midrule
        TF-C          & \pmstd{59.28}{0.86} & \pmstd{77.66}{0.64} & \pmstd{47.79}{2.35} & \pmstd{91.89}{2.02} & \pmstd{\mathbf{24.16}}{0.86} & \pmstd{\underline{61.69}}{0.65} & \underline{77.08} \\
        MOMENT-Large  & \pmstd{58.73}{1.16} & \pmstd{78.45}{0.55} & \pmstd{49.60}{0.55} & \pmstd{88.55}{0.64} & \pmstd{17.30}{0.44} & \pmstd{53.87}{0.49} & 72.75 \\
        TimeMoE-Large & \pmstd{59.69}{0.49} & \pmstd{76.95}{0.27} & \pmstd{37.53}{1.17} & \pmstd{85.94}{0.75} & \pmstd{20.51}{0.16} & \pmstd{59.29}{0.57} & 74.93 \\
        TiRex         & \pmstd{58.57}{0.72} & \pmstd{\underline{79.56}}{0.73} & \pmstd{42.89}{0.95} & \pmstd{89.52}{1.05} & \pmstd{19.05}{0.75} & \pmstd{55.25}{0.90} & 74.78 \\
        Sundial       & \pmstd{58.88}{1.09} & \pmstd{73.47}{1.21} & \pmstd{48.69}{0.40} & \pmstd{88.12}{0.81} & \pmstd{20.54}{0.37} & \pmstd{59.80}{0.52} & 73.80 \\
        Chronos2-Base & \pmstd{59.61}{0.69} & \pmstd{78.90}{0.69} & \pmstd{48.77}{0.42} & \pmstd{\underline{92.63}}{0.50} & \pmstd{17.92}{0.91} & \pmstd{53.01}{0.49} & 74.85 \\
        NormWear      & \pmstd{59.50}{0.72} & \pmstd{75.85}{0.57} & \pmstd{43.03}{0.16} & \pmstd{91.92}{0.98} & \pmstd{20.10}{0.78} & \pmstd{61.22}{0.79} & 76.34 \\
        MIRA-Large    & \pmstd{\underline{59.95}}{0.92} & \pmstd{78.71}{1.13} & \pmstd{\underline{50.09}}{1.22} & \pmstd{90.07}{0.81} & \pmstd{\underline{23.52}}{0.22} & \pmstd{59.94}{1.20} & 76.24 \\
        \midrule
        SOTER         & \pmstd{\mathbf{60.83}}{0.82} & \pmstd{\mathbf{81.70}}{0.65} & \pmstd{\mathbf{50.90}}{0.36} & \pmstd{\mathbf{95.62}}{0.12} & \pmstd{23.46}{0.68} & \pmstd{\mathbf{65.04}}{0.68} & $\mathbf{80.79}$ \\
        \bottomrule
    \end{tabular}
\end{table*}

\begin{figure*}[t]
    \centering
    \includegraphics[width=\textwidth]{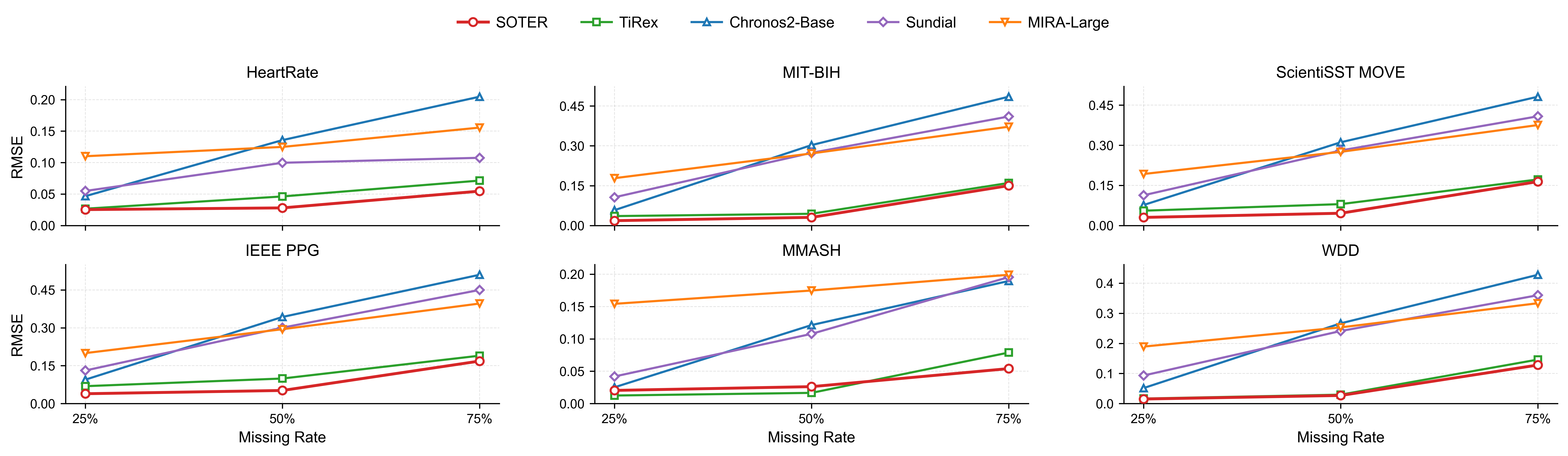}
    \caption{Continuous-time imputation under increasing missing rates.
    RMSE on six OOD wearable datasets at 25\%, 50\%, and 75\% missing
    rates under the identical causal-prefix protocol. Values are means
    over three runs.}
    \label{fig:imputation}
\end{figure*}

\subsection{Zero-Shot Future Forecasting}
\label{sec:forecasting}
We evaluate zero-shot future forecasting on six out-of-distribution
(OOD) wearable physiological datasets, reporting RMSE and MAE in the
normalized space (Table~\ref{tab:main_results}). SOTER achieves the best
RMSE on four of the six datasets and the best MAE on five, while
activating only $20.29$M parameters per inference step. 

MMASH is the only dataset on which SOTER achieves neither the best RMSE
nor the best MAE. To provide additional context, its normalized
permutation entropy (NPE) is $0.334$, the lowest among the six evaluated
datasets and substantially below the $0.878$ measured on MIT-BIH. This
difference indicates that MMASH exhibits markedly lower temporal
complexity than the other evaluation datasets. We hypothesize that this
property contributes to SOTER's relative disadvantage on MMASH. Further
dataset statistics and complexity analyses are provided in the
Appendix~\ref{app:datasets}. Despite this exception, SOTER
achieves the best performance on nine of the twelve dataset--metric
combinations, demonstrating strong OOD transfer for zero-shot
physiological forecasting.

\subsection{Downstream Task Evaluation}
\label{sec:downstream_tasks}

We further evaluate the same frozen SOTER on classification and
continuous-time imputation.

\paragraph{Classification.}
We adopt a frozen linear-probe protocol: the backbone remains fixed, and
only a logistic-regression classifier is trained on top of its
representations. This protocol provides a controlled measure of the
discriminative information retained by the pre-trained model. Evaluation
follows the LOSO split, and we report the mean and standard deviation
over three runs. Because the evaluated datasets exhibit class imbalance,
we report Macro-F1 and Macro-AUROC rather than accuracy, and use the
average Macro-AUROC across datasets as the overall metric. As shown in Table~\ref{tab:classification}, SOTER achieves the highest
average Macro-AUROC of $80.79$, outperforming the representation-learning
baseline TF-C ($77.08$) and the physiological foundation models NormWear
($76.34$) and MIRA ($76.24$). It also obtains the best Macro-AUROC on all
three datasets. On MMASH, whose six-class distribution is the most
imbalanced among the evaluated classification datasets, all methods
achieve relatively low Macro-F1. SOTER trails TF-C on this metric but
retains the highest Macro-AUROC, with all evaluated methods remaining
above the random baseline in Macro-AUROC. Appendix~\ref{app:label_efficiency} further shows that SOTER retains the strongest frozen linear-probe performance on ScientiSST MOVE across label budgets ranging from 1\% to 100\%.

\paragraph{Continuous-Time Imputation.}
Wearable signals are acquired and consumed as streams, where missing
values must be estimated online using past observations alone. We
therefore consider online causal imputation rather than offline
bidirectional reconstruction. For each missing point, only its causal
prefix is provided as context, and the CDE/ODE decoder evolves the latent
state to the corresponding timestamp without write-back or autoregressive
error accumulation. All models are evaluated under the same causal-prefix protocol, including
identical missingness construction, evaluation points, and metrics. We
consider missing rates of $25\%$, $50\%$, and $75\%$. Under this protocol,
SOTER transfers to imputation zero-shot, without task-specific
optimization. As shown in Figure~\ref{fig:imputation}, SOTER achieves the lowest RMSE on most datasets at each evaluated missing rate. At the highest missing rate of $75\%$, SOTER ranks first on all six datasets. TiRex outperforms
SOTER on MMASH at the two lower missing rates. As missingness increases, Chronos-2, Sundial, and MIRA exhibit a more
pronounced degradation, whereas SOTER and TiRex remain comparatively
stable. This trend is consistent with the benefit of explicitly evolving
the latent state toward arbitrary query timestamps under sparse causal
observations. Full numerical results and standard deviations are provided
in the Appendix~\ref{app:imputation_results}.

\subsection{Ablation Analysis}
\label{sec:ablation}

\begin{table}[t!]
    \caption{Ablation study of SOTER on one single-channel and two
    multi-channel datasets. Entries are rescaled by $10^{2}$, as in
    Table~\ref{tab:main_results}. Bold indicates the best result in each
    column.}
    \label{tab:ablation}
    \centering
    \small
    \renewcommand{\arraystretch}{0.8}
    \setlength{\tabcolsep}{0.55mm}
    \resizebox{\columnwidth}{!}{%
    \begin{tabular}{l cc cc cc}
        \toprule
        \multirow{2}{*}{Method}
        & \multicolumn{2}{c}{\shortstack{HeartRate\\(SC)}}
        & \multicolumn{2}{c}{\shortstack{MIT-BIH\\(MC)}}
        & \multicolumn{2}{c}{\shortstack{ScientiSST\\MOVE (MC)}} \\
        \cmidrule(lr){2-3}
        \cmidrule(lr){4-5}
        \cmidrule(lr){6-7}
        & RMSE & MAE & RMSE & MAE & RMSE & MAE \\
        \midrule
        \rowcolor[HTML]{E5E5E5}
        SOTER                & $\mathbf{1.41}$ & $\mathbf{0.95}$ & $\mathbf{1.29}$ & $\mathbf{0.62}$ & $\mathbf{2.26}$ & $\mathbf{1.17}$ \\
        \quad w/o CD Layer   & 1.43 & 0.98 & 2.14 & 1.68 & 3.78 & 2.15 \\
        \quad w/o PSD-MoE    & 1.46 & 1.01 & 1.35 & 0.73 & 2.36 & 1.22 \\
        \quad w/o Neural CDE & 1.54 & 1.09 & 1.67 & 0.79 & 2.64 & 1.39 \\
        \midrule
        \quad CD Layer@$5$th & 1.42 & 0.97 & 1.47 & 0.82 & 2.32 & 1.25 \\
        \quad CD Layer@$4$th & 1.43 & 0.99 & 1.73 & 1.04 & 2.47 & 1.39 \\
        \quad CD Layer@$3$rd & 1.43 & 1.01 & 1.98 & 1.31 & 2.72 & 1.63 \\
        \bottomrule
    \end{tabular}%
    }
\end{table}

We ablate the three core components of SOTER on the single-channel
HeartRate dataset and the multi-channel MIT-BIH and ScientiSST MOVE
datasets. In Table~\ref{tab:ablation}, w/o CD Layer replaces the
channel-dependent layer with channel-independent processing;
w/o PSD-MoE replaces the deterministic spectral router with a
learned router trained using an auxiliary load-balancing loss; and
w/o Neural CDE reduces the continuous-time decoder to a pure
Neural ODE. Because the CDE control path is activated only in the
missingness branch of pre-training, the final variant also removes this
branch and therefore measures the combined contribution of the CDE
mechanism and missingness-aware pre-training.

Removing any of the three components increases both RMSE and MAE on all
evaluated datasets. Removing the CD layer has little effect on
single-channel HeartRate but causes substantially larger degradation on
the two multi-channel datasets, supporting its role in modeling
cross-channel dependencies. Removing the Neural CDE also increases the
error on HeartRate, showing that its benefit is not limited to
multi-channel inputs and is complementary to cross-channel modeling.

Replacing PSD-MoE with the learned auxiliary-loss-based router consistently yields higher errors. Under the
same training configuration, the deterministic spectral router also
introduces no learned routing parameters and reduces the average
pre-training time from \textbf{2.45} to \textbf{2.17} seconds per step. Finally, we
move the CD layer from the default top block to progressively shallower
positions, where CD Layer@$k$th denotes placement at the $k$-th of the
six backbone layers. Errors increase monotonically on both multi-channel
datasets as the CD layer is moved downward, whereas HeartRate remains
nearly unchanged. This trend suggests that, within SOTER, cross-channel
interaction is most effective after the channel-specific temporal
representations have been sufficiently refined.

\begin{table}[t!]
    \caption{Corpus-matched comparison on zero-shot forecasting.
    SOTER and the two corpus-matched MIRA-Large variants are trained for
    one full epoch on the same physiological corpus under the same
    training budget. Results are reported as the mean and standard
    deviation over four input/output window pairs and entries are rescaled by
    $10^{2}$, as in Table~\ref{tab:main_results}. Lower values are
    better.}
    \label{tab:corpus_matched}
    \centering
    \small
    \renewcommand{\arraystretch}{0.92}
    \setlength{\tabcolsep}{2.2pt}
    \resizebox{\columnwidth}{!}{%
    \begin{tabular}{lcccc}
        \toprule
        \multirow{2}{*}{Method}
        & \multicolumn{2}{c}{MIT-BIH}
        & \multicolumn{2}{c}{ScientiSST MOVE} \\
        \cmidrule(lr){2-3}
        \cmidrule(lr){4-5}
        & RMSE & MAE & RMSE & MAE \\
        \midrule
        \rowcolor[HTML]{E5E5E5}
        SOTER                  & \pmstd{\mathbf{1.29}}{0.22} & \pmstd{\mathbf{0.62}}{0.10} & \pmstd{\mathbf{2.26}}{0.12} & \pmstd{\mathbf{1.17}}{0.04} \\
        MIRA-Large             & \pmstd{2.20}{0.14} & \pmstd{1.54}{0.12} & \pmstd{4.76}{0.16} & \pmstd{3.73}{0.12} \\
        MIRA-Large (scratch)   & \pmstd{2.03}{0.17} & \pmstd{1.38}{0.11} & \pmstd{3.85}{0.20} & \pmstd{2.91}{0.18} \\
        MIRA-Large (continued) & \pmstd{1.79}{0.19} & \pmstd{1.08}{0.17} & \pmstd{3.12}{0.19} & \pmstd{2.37}{0.14} \\
        \bottomrule
    \end{tabular}%
    }
\end{table}

\subsection{Corpus-Matched Comparison}
\label{sec:corpus_matched}
The remaining question is whether the pre-training corpus alone explains
this advantage. We therefore pre-train MIRA-Large on the same
physiological corpus under the same one-epoch training budget and
optimization schedule, both from scratch and by continued pre-training.
Here, MIRA-Large denotes the officially released checkpoint,
MIRA-Large (scratch) is randomly initialized and trained on our
corpus, and MIRA-Large (continued) is initialized from the
official checkpoint and further pre-trained on the same corpus. All
models are evaluated using the same zero-shot forecasting protocol and
input/output window settings. The results are reported in Table~\ref{tab:corpus_matched}. Training MIRA-Large on the physiological corpus improves upon its
official checkpoint, with continued pre-training producing the strongest
MIRA variant on both datasets. Nevertheless, SOTER remains ahead on all
four metrics, including reductions over MIRA-Large (continued) from
$1.79$ to $1.29$ RMSE on MIT-BIH and from $3.12$ to $2.26$ on
ScientiSST MOVE. These results indicate that corpus matching contributes
to performance, but the corpus alone does not account for SOTER's gains
on the evaluated datasets.

\subsection{Interpretability of PSD-Guided Routing}
\label{sec:routing_interpretability}

\begin{figure}[t!]
    \centering
    \includegraphics[width=0.75\columnwidth]{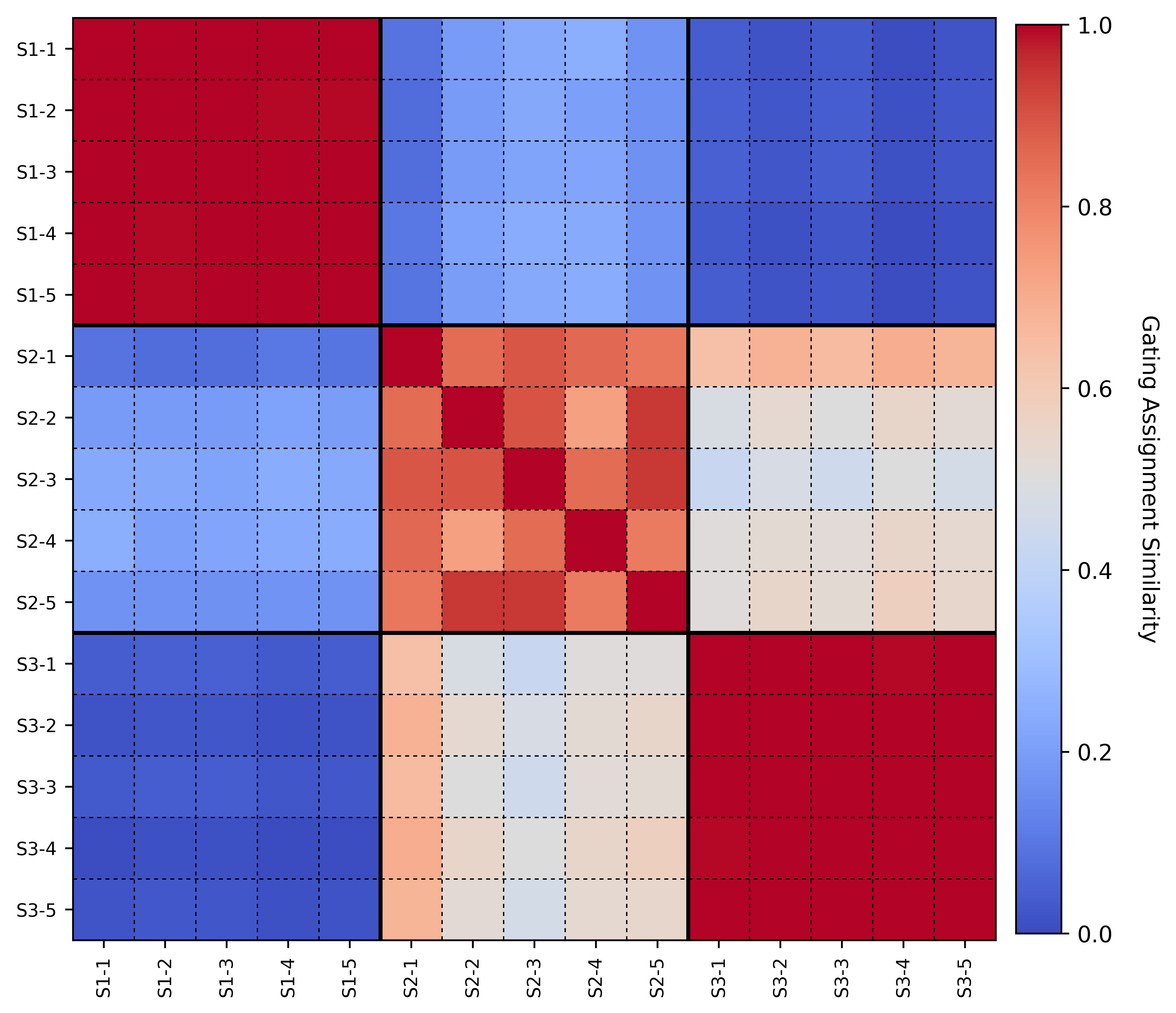}
    \caption{Pairwise cosine similarity between gate-assignment vectors
    for three signal types (S1--S3), with five segments sampled from each
    signal using seed 42. The block-diagonal structure indicates that
    signals from different frequency regimes receive separable routing
    assignments.}
    \label{fig:gate_sim}
\end{figure}

To examine whether the spectral statistic used by the router reflects
the frequency content of the underlying signal, we analyze three signals
from distinct regimes: high-frequency ECG from MIT-BIH (S1),
mid-frequency PPG from IEEE PPG (S2), and low-frequency heart rate from
HeartRate (S3). We sample five segments from each signal, pass them
through SOTER, and record the gate-assignment vector at the topmost
PSD-guided MoE layer. Because the router operates on latent
representations rather than raw inputs, the spectral structure available
after the attention layers need not directly preserve the input
frequency characteristics.

Figure~\ref{fig:gate_sim} nevertheless exhibits clear within-signal
similarity and cross-signal separation. Segments from the same signal
form high-similarity blocks, S1 and S3 are well separated, and S2 lies
between these two regimes. This pattern indicates that the latent
spectral statistic remains associated with the spectral characteristics of the underlying signals, providing an explicit and inspectable basis for expert routing.

\subsection{Noise Robustness Analysis}
\label{sec:noise_robustness}

\begin{figure}[t]
    \centering
    \includegraphics[width=\columnwidth]{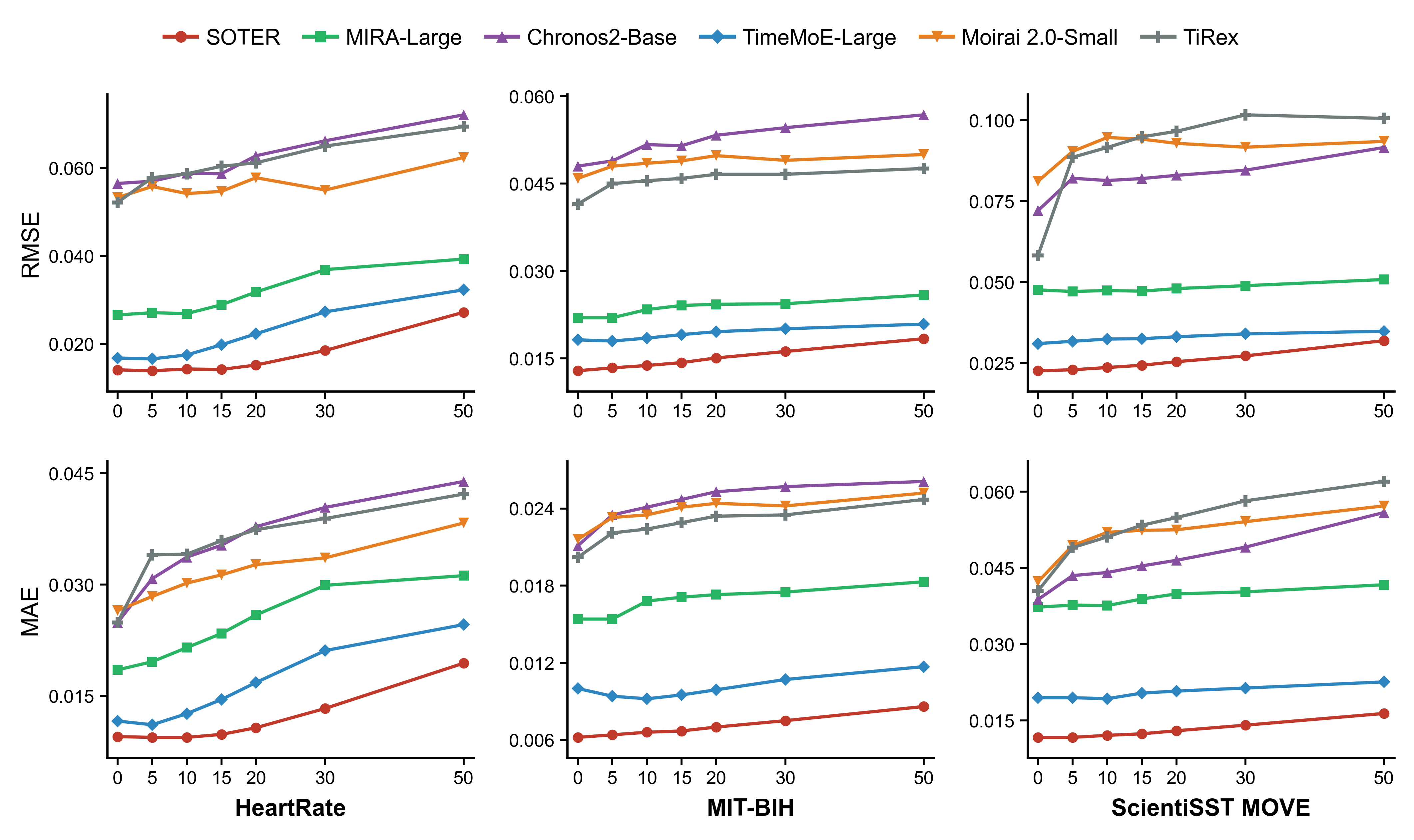}
    \caption{Zero-shot forecasting under additive Gaussian noise.
    Noise with standard deviation
    $\beta\sigma_{\mathrm{signal}}$ is injected into the input context,
    where $\sigma_{\mathrm{signal}}$ is the per-channel signal scale
    estimated from the training split and
    $\beta\in\{0.05,\ldots,0.5\}$. RMSE and MAE are computed against the
    clean future on three datasets.}
    \label{fig:noise_robustness}
\end{figure}

To emulate acquisition noise in real-world wearable sensing, we inject
scale-proportional additive Gaussian noise into the observed context and
evaluate zero-shot prediction against the clean future
(Figure~\ref{fig:noise_robustness}). Across all six noise levels, both
metrics, and all three datasets, SOTER achieves the lowest prediction
error among the evaluated methods. Even under the strongest corruption
level, $\beta=0.5$, its performance remains comparable to or better than
that of the baselines evaluated on clean inputs ($\beta=0$), demonstrating
strong robustness to additive acquisition noise.

\begin{table}[t!]
    \caption{OOD forecasting performance at different levels of
    cumulative pre-training exposure. Results are the mean and standard
    deviation over four input/output windows and are rescaled by
    $10^{2}$. Lower values are better.}
    \label{tab:pretraining_exposure}
    \centering
    \scriptsize
    \renewcommand{\arraystretch}{1.05}
    \setlength{\tabcolsep}{2.25pt}
    \resizebox{\columnwidth}{!}{%
    \begin{tabular}{lcccccc}
        \toprule
        \multirow{2}{*}{Exposure}
        & \multicolumn{2}{c}{MIT-BIH}
        & \multicolumn{2}{c}{\shortstack{ScientiSST MOVE}}
        & \multicolumn{2}{c}{WDD} \\
        \cmidrule(lr){2-3}
        \cmidrule(lr){4-5}
        \cmidrule(lr){6-7}
        & RMSE & MAE & RMSE & MAE & RMSE & MAE \\
        \midrule
        $10\%$  & \pmstd{1.80}{0.20} & \pmstd{1.35}{0.34} & \pmstd{3.33}{0.30} & \pmstd{2.21}{0.31} & \pmstd{4.89}{0.16} & \pmstd{3.41}{0.26} \\
        $25\%$  & \pmstd{1.52}{0.15} & \pmstd{1.05}{0.13} & \pmstd{2.59}{0.10} & \pmstd{1.80}{0.12} & \pmstd{3.16}{0.25} & \pmstd{2.19}{0.41} \\
        $50\%$  & \pmstd{1.32}{0.23} & \pmstd{0.88}{0.07} & \pmstd{2.44}{0.14} & \pmstd{1.39}{0.15} & \pmstd{2.69}{0.19} & \pmstd{1.47}{0.13} \\
        $75\%$  & \pmstd{\underline{1.29}}{0.25} & \pmstd{\underline{0.63}}{0.09} & \pmstd{\underline{2.32}}{0.15} & \pmstd{\underline{1.25}}{0.08} & \pmstd{\underline{2.22}}{0.19} & \pmstd{\underline{1.17}}{0.12} \\
        \rowcolor[HTML]{E5E5E5}
        $100\%$ & \pmstd{\mathbf{1.29}}{0.22} & \pmstd{\mathbf{0.62}}{0.10} & \pmstd{\mathbf{2.26}}{0.12} & \pmstd{\mathbf{1.17}}{0.04} & \pmstd{\mathbf{2.19}}{0.19} & \pmstd{\mathbf{1.05}}{0.05} \\
        \bottomrule
    \end{tabular}%
    }
\end{table}

\subsection{Cumulative Pre-training Exposure Analysis}
\label{sec:pretraining_exposure}
To examine how OOD transfer evolves over the single-epoch pre-training
run, we evaluate checkpoints corresponding to cumulative corpus
exposures of $10\%$, $25\%$, $50\%$, $75\%$, and $100\%$ under the same
four forecasting-window settings, as shown in Table~\ref{tab:pretraining_exposure}. Across all three OOD datasets, increasing cumulative pre-training
exposure consistently reduces both RMSE and MAE, indicating progressively
improved downstream transfer over the course of training. The rate of
saturation, however, is dataset- and metric-dependent. On MIT-BIH, the
RMSE at $50\%$ exposure is already within $2.3\%$ of the final result,
whereas the MAE approaches its final value only after $75\%$ exposure.
On ScientiSST MOVE, the RMSE and MAE at $50\%$ exposure remain $8.0\%$
and $18.8\%$ above their final values, respectively; these gaps narrow
to $2.7\%$ and $6.8\%$ at $75\%$. WDD exhibits the strongest dependence
on later-stage pre-training, particularly in MAE. Overall, OOD transfer
improves consistently as pre-training progresses, while the marginal
performance gains gradually diminish at dataset- and metric-dependent
rates.

\section{Conclusion}
\label{sec:conclusion}
We presented SOTER, a generative time-series foundation model that jointly models
cross-channel coupling, spectrum-guided expert specialization, and
continuous-time dynamics for wearable physiological signals. Across
out-of-distribution forecasting, linear-probe classification, and causal
continuous-time imputation, a single pre-trained backbone achieves strong
and consistent transfer while activating only $20.29$M parameters per
inference step, and remains robust under substantial missingness and
additive acquisition noise. Component ablations support the
non-redundant roles of the three modeling mechanisms, while the
full-budget corpus-matched comparison shows that physiological
pre-training improves MIRA-Large but does not alone account for SOTER's
advantage. These findings suggest that aligning model structure with the
channel organization, spectral heterogeneity, and continuous-time nature
of physiological processes is an effective direction for
domain-specialized time-series foundation models. Future work will extend
this framework to broader physiological modalities and populations,
larger-scale pre-training, and prospective clinical evaluation.

\bibliographystyle{ACM-Reference-Format}
\bibliography{sample-base}

\appendix

\section{Datasets and Statistics}
\label{app:datasets}

We briefly describe the datasets used in this work. SOTER is pre-trained
on a wearable physiological corpus containing approximately 226 billion
time points, assembled from five publicly available and de-identified
datasets: MIMIC-III Waveform
~\cite{mimic3wdb,johnson2016mimic},
Sleep-EDF~\cite{sleepedf},
PTB-XL~\cite{wagner2020ptb,PhysioNet-ptb-xl-1.0.3},
WESAD~\cite{wesad}, and
Chapman-ECG
~\cite{chapmanecg,PhysioNet-chapmanecg-arrhythmia-1.0.0}.
Evaluation is conducted on six publicly available physiological
datasets recorded using wearable or ambulatory sensing devices:
HeartRate~\cite{heartrate},
MIT-BIH~\cite{mitbih},
ScientiSST MOVE~\cite{scientisst-move1},
IEEE PPG~\cite{ieeeppg},
MMASH~\cite{PhysioNet-mmash,mmash}, and
WDD~\cite{PhysioNet-wdd,wdd}.
We introduce the pre-training corpora first, followed by the evaluation
datasets.

\subsection{Pre-training Corpora}

\paragraph{MIMIC-III Waveform}
The MIMIC-III Waveform Database comprises $67{,}830$ record sets from
approximately $30{,}000$ intensive-care-unit patients. Most record sets
pair digitized physiological waveforms---typically ECG, arterial blood
pressure, respiration, and PPG---with periodically sampled numerical
vital signs, forming quasi-continuous patient recordings spanning from
days to several weeks. A subset is matched and time-aligned with the
MIMIC-III Clinical Database.

\paragraph{Sleep-EDF}
Sleep-EDF contains $197$ whole-night polysomnographic recordings
comprising EEG, EOG, chin EMG, and event markers, with some recordings
additionally providing respiration and body temperature. Each recording
is accompanied by an expert-scored hypnogram annotated according to the
Rechtschaffen and Kales standard.

\paragraph{PTB-XL}
PTB-XL is a large clinical electrocardiography dataset containing
$21{,}799$ ten-second, 12-lead ECG recordings from $18{,}869$ patients.
Each record was annotated by up to two cardiologists with potentially
multiple statements drawn from $71$ SCP-ECG-conformant diagnostic, form,
and rhythm categories, together with demographic and signal-property
metadata.

\paragraph{WESAD}
WESAD is a multimodal wearable dataset for stress and affect detection,
recorded from wrist- and chest-worn devices on $15$ subjects. It includes
blood volume pulse, ECG, electrodermal activity, electromyography,
respiration, body temperature, and three-axis acceleration, and covers
three affective states: neutral, stress, and amusement.

\paragraph{Chapman-ECG}
The Chapman-ECG database provides 12-lead ECG recordings from
$45{,}152$ patients sampled at $500$~Hz. It covers multiple common
cardiac rhythms and additional cardiovascular conditions, with labels
provided by professional experts, and supports the development and
evaluation of methods for arrhythmia and related cardiovascular
analysis.

\subsection{Evaluation Datasets}

\paragraph{HeartRate}
Drawn from the Monash Time Series Extrinsic Regression Archive,
HeartRate pairs short photoplethysmography (PPG) segments with continuous
heart-rate targets. It serves as the single-channel physiological
benchmark in our evaluation.

\paragraph{MIT-BIH}
The MIT-BIH Arrhythmia Database contains $48$ half-hour, two-channel
ambulatory ECG excerpts from $47$ subjects, digitized at $360$~Hz per
channel with 11-bit resolution over a $10$~mV range. Each beat was
independently annotated by two or more cardiologists, yielding
approximately $110{,}000$ reference annotations.

\paragraph{ScientiSST MOVE}
ScientiSST MOVE records everyday activities---including lifting a chair,
greeting, gesticulating, walking, and running---from $17$ healthy
volunteers using a chestband, an armband, and an Empatica~E4 wristband.
The dataset includes multi-channel EDA, PPG, and ECG signals, together
with bicep EMG, wrist temperature, and actigraphy, providing
physiological measurements under dynamic and less-controlled conditions.

\paragraph{IEEE PPG}
Originating from the IEEE Signal Processing Cup 2015, IEEE PPG targets
heart-rate estimation during physical exercise. It contains $3{,}096$
multi-channel sequences comprising two-channel wrist PPG, three-axis
acceleration, and one-channel chest ECG, recorded simultaneously at
$125$~Hz from subjects aged 18--35.

\paragraph{MMASH}
MMASH provides 24-hour recordings of continuous beat-to-beat heart data,
triaxial acceleration, sleep quality, physical activity, and
psychological characteristics from $22$ healthy participants, together
with saliva biomarkers and activity logs. It supports analysis of the
relationships among physical activity, sleep, and physiological state.

\paragraph{WDD}
WDD records physiological signals during structured acute-stress
induction and aerobic or anaerobic exercise using the Empatica~E4
wearable device. It includes blood volume pulse, motion-based activity,
skin temperature, and electrodermal activity. The dataset comprises
$36$ volunteers for stress sessions, $30$ for aerobic exercise, and
$31$ for anaerobic exercise.

Table~\ref{tab:eval-stats} summarizes the channel configuration and
normalized permutation entropy (NPE) of each evaluation dataset. We use
NPE as a scalar indicator of temporal complexity, with larger values
corresponding to less predictable dynamics. MMASH has the lowest NPE
among the six datasets ($0.334$), whereas MIT-BIH has the highest
($0.878$), providing additional context for the forecasting results
discussed in the main paper. All datasets used in this study are \textbf{publicly
available and de-identified}, and no personally identifiable information
was accessed.

\begin{table}[t!]
    \caption{Type, number of feature channels, and normalized permutation
    entropy (NPE) of the six evaluation datasets. SC and MC denote
    single-channel and multi-channel datasets, respectively. Higher NPE
    indicates less predictable temporal dynamics.}
    \label{tab:eval-stats}
    \centering
    \begin{tabular}{lccc}
        \toprule
        Dataset & Type & \#Channels & NPE \\
        \midrule
        HeartRate       & SC & 1 & 0.469 \\
        MIT-BIH         & MC & 2 & 0.878 \\
        ScientiSST MOVE & MC & 5 & 0.643 \\
        IEEE PPG        & MC & 5 & 0.646 \\
        MMASH           & MC & 6 & 0.334 \\
        WDD             & MC & 7 & 0.618 \\
        \bottomrule
    \end{tabular}
\end{table}

\section{Pre-training Setup and Logs}
\label{app:pretraining}

This section details the pre-training configuration of SOTER in
sufficient depth to support reproduction and visualizes its convergence.

\paragraph{Compute and data pipeline.}
SOTER is pre-trained for a single epoch on the full corpus of
approximately 226 billion time points, amounting to $2{,}076{,}572$
optimization steps, on $8\times$ NVIDIA H200 GPUs. We use a per-GPU
batch size of $8$ with no gradient accumulation, resulting in a global
batch size of $64$. The maximum sequence length is $512$, and data
loading uses $12$ workers. Training is conducted in bf16 mixed precision,
and attention is computed using an eager, non-fused implementation. The
windowed-sample cache is pre-warmed before optimization begins.

\paragraph{Optimization.}
We optimize SOTER using AdamW with a learning rate of
$3\times10^{-4}$, $\beta=(0.9,0.95)$, $\epsilon=10^{-8}$, and weight
decay of $0.01$. We employ a cosine-annealing schedule with linear
warmup. The warmup is deliberately short
($warmup\_steps=20$), because the single-epoch run contains a
large number of optimization steps, and the cosine schedule spans the
actual number of steps in one epoch. No gradient clipping is applied,
and all runs use seed $42$.

\paragraph{Decoder solver and PSD routing.}
The continuous-time decoder is integrated using the Dormand--Prince 5(4) solver (dopri5) with a tolerance of $10^{-5}$. The PSD-guided
router is implemented using a parallel, strictly causal prefix discrete
Fourier transform (DFT). The effects of the DFT implementation, solver
family, and solver tolerance are examined in
Appendix~\ref{app:sensitivity}.

\begin{figure}[t!]
    \centering
    \includegraphics[width=\columnwidth]
    {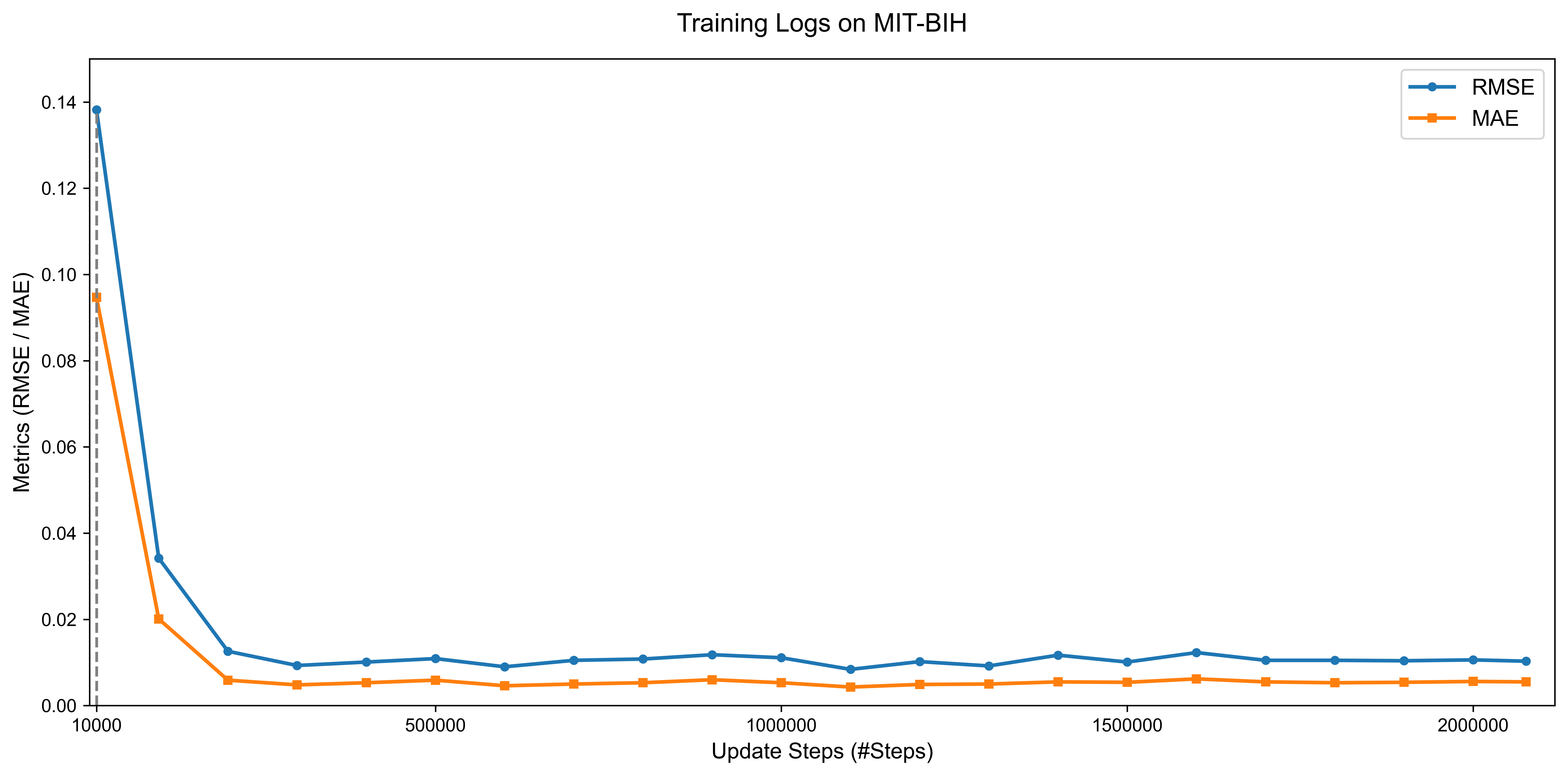}
    \caption{Pre-training convergence on the MIT-BIH
    out-of-distribution quick evaluation. Every $100{,}000$ optimization
    steps, RMSE and MAE in the normalized space are computed on $32$
    held-out MIT-BIH samples using input/output windows of $128/64$.
    Both metrics decrease sharply during the early evaluation intervals
    and subsequently remain on a low and stable plateau throughout the
    remainder of the single-epoch run.}
    \label{fig:pretrain-log}
\end{figure}

\paragraph{Convergence monitoring.}
Every 100,000 optimization steps, we conduct a lightweight
out-of-distribution evaluation on MIT-BIH. We evaluate 32 held-out
samples using input and output window lengths of 128 and 64,
respectively, with seed 123. RMSE and MAE are reported in the
normalized space, where the per-channel MinMax normalization parameters
are estimated exclusively from the MIT-BIH training split.
Figure~\ref{fig:pretrain-log} visualizes the resulting trajectory.

As shown in Figure~\ref{fig:pretrain-log}, both RMSE and MAE decrease
rapidly during the earliest evaluation intervals and subsequently
stabilize at low values for the remainder of the single-epoch run,
indicating stable pre-training convergence.

\begin{table*}[t!]
    \caption{Sensitivity analysis of the DFT implementation,
    continuous-time solver, and solver tolerance. Each configuration is
    evaluated in a separate $100{,}000$-step sensitivity run, independent
    of full pre-training, on zero-shot forecasting over MIT-BIH using
    $256$ samples, input/output windows of $128/64$, and seed $42$. RMSE
    and MAE are reported in the normalized space, and latency denotes the
    average optimization time per step within this sensitivity run. Bold
    and underlined values indicate the best and second-best results in
    each metric, respectively. The highlighted row is the configuration
    adopted by SOTER.}
    \label{tab:sensitivity}
    \centering
    \renewcommand{\arraystretch}{0.92}
    \setlength{\tabcolsep}{10pt}
    \begin{tabular}{llcccc}
        \toprule
        DFT Implementation & Solver & Tolerance & RMSE $\downarrow$ & MAE $\downarrow$ & \shortstack{Latency (s/step) $\downarrow$} \\
        \midrule
        For Loop & dopri5 & $10^{-6}$ & 0.0477 & 0.0350 & 9.88 \\
        For Loop & dopri5 & $10^{-5}$ & $\mathbf{0.0470}$ & $\mathbf{0.0345}$ & 9.92 \\
        For Loop & dopri5 & $10^{-4}$ & 0.0487 & 0.0359 & 9.90 \\
        For Loop & dopri5 & $10^{-3}$ & 0.0504 & 0.0373 & 9.80 \\
        \midrule
        For Loop & rk4 & $10^{-6}$ & 0.0482 & 0.0353 & 9.79 \\
        For Loop & rk4 & $10^{-5}$ & 0.0484 & 0.0355 & 9.04 \\
        For Loop & rk4 & $10^{-4}$ & 0.0487 & 0.0358 & 9.05 \\
        For Loop & rk4 & $10^{-3}$ & 0.0488 & 0.0359 & 9.08 \\
        \midrule
        Parallel Prefix & dopri5 & $10^{-6}$ & 0.0483 & 0.0353 & 2.18 \\
        \rowcolor[HTML]{FEEAD2}
        Parallel Prefix & dopri5 & $10^{-5}$ & $\underline{0.0473}$ & $\underline{0.0347}$ & 2.17 \\
        Parallel Prefix & dopri5 & $10^{-4}$ & 0.0478 & 0.0350 & 2.09 \\
        Parallel Prefix & dopri5 & $10^{-3}$ & 0.0488 & 0.0359 & $\underline{2.08}$ \\
        \midrule
        Parallel Prefix & rk4 & $10^{-6}$ & 0.0479 & 0.0352 & 2.14 \\
        Parallel Prefix & rk4 & $10^{-5}$ & 0.0481 & 0.0355 & 2.12 \\
        Parallel Prefix & rk4 & $10^{-4}$ & 0.0482 & 0.0356 & 2.17 \\
        Parallel Prefix & rk4 & $10^{-3}$ & 0.0479 & 0.0351 & $\mathbf{2.03}$ \\
        \bottomrule
    \end{tabular}
\end{table*}

\section{Key Sensitivity Analysis}
\label{app:sensitivity}
To balance predictive performance and training efficiency, we analyze
the effects of the discrete Fourier transform (DFT) implementation,
continuous-time solver, and solver tolerance. We compare a parallel-prefix
implementation with a loop-based implementation for the causal prefix
DFT, consider dopri5 and rk4 as the numerical solvers,
and evaluate tolerances of $10^{-6}$, $10^{-5}$, $10^{-4}$, and
$10^{-3}$.

All configurations are evaluated after $100{,}000$ optimization steps
using zero-shot forecasting on MIT-BIH with $256$ samples, input/output
windows of $128/64$, and random seed $42$. We report RMSE and MAE in the
normalized space, together with the average latency per optimization
step. Table~\ref{tab:sensitivity} summarizes the results.

The loop-based DFT consistently incurs substantially higher latency than
the parallel-prefix implementation. Although the lowest forecasting
errors are obtained by the loop-based DFT with dopri5 and a
tolerance of $10^{-5}$, this configuration requires approximately four
times the per-step latency of its parallel-prefix counterpart. The
fastest configuration combines the parallel-prefix DFT, rk4,
and a tolerance of $10^{-3}$, but yields slightly higher prediction
errors. We therefore adopt the parallel-prefix DFT with dopri5
and a tolerance of $10^{-5}$, which adds only $0.14$ seconds per step
relative to the fastest configuration while providing a more favorable
accuracy--efficiency trade-off. This choice is consistent with the
pre-training configuration described in
Appendix~\ref{app:pretraining}.

\section{From Neural CDE to Neural ODE}
\label{app:cde_ode}

This section formalizes the relationship between the two operating
regimes of SOTER's terminal continuous-time decoder. The Neural ODE used
for future forecasting is not implemented as a separate module; rather,
it is obtained from the observation-guided Neural CDE regime by setting
the external control path to zero. Both regimes share the same vector
field and parameter set $\theta$. The resulting reduction is an
algebraic identity rather than a numerical approximation.

The notation follows the main paper. The terminal backbone state at the
last observed timestamp $t_T$ is denoted by $h_T$, and $t_{T+1}$ denotes
the queried timestamp. The formulation below provides the
implementation-level realization of
Eqs.~\eqref{eq:cde_dynamics} and~\eqref{eq:integration}.

\subsection{Setup and Notation}
\label{app:cde_ode_setup}

The autoregressive backbone produces the terminal hidden state
\begin{equation}
    h_T = h(t_T) \in \mathbb{R}^{H},
    \label{eq:app_terminal_state}
\end{equation}
where $H$ denotes the hidden dimension. The continuous-time decoder
propagates $h_T$ toward the queried timestamp $t_{T+1}$.

We define the integration interval as
\begin{equation}
\begin{aligned}
    \delta t &= t_{T+1}-t_T, \\
    \Delta t &= \max(\delta t,\varepsilon),
    \qquad \varepsilon=10^{-5},
\end{aligned}
\label{eq:app_delta_t}
\end{equation}
where the lower bound $\varepsilon$ is a numerical safeguard against a
degenerate zero-length interval. For all non-degenerate queries,
$\Delta t=\delta t$.

\paragraph{Normalized integration variable.}
Numerical integration is performed over a normalized variable
$s\in[0,1]$. Physical time is related to $s$ through the affine
reparameterization
\begin{equation}
\begin{aligned}
    t(s) &= t_T+s\Delta t, \qquad s\in[0,1], \\
    \frac{\mathrm{d}t}{\mathrm{d}s} &= \Delta t.
\end{aligned}
\label{eq:app_reparameterization}
\end{equation}
Consequently, $s=0$ corresponds to the last observed timestamp $t_T$,
and $s=1$ corresponds to the queried endpoint.

\paragraph{Vector field.}
The decoder dynamics are generated by a single multilayer perceptron
$f_{\theta}$ with parameters $\theta$:
\begin{equation}
\begin{aligned}
    f_{\theta}
    &\colon
    [0,1]\times\mathbb{R}^{H}\times\mathbb{R}^{d_z}
    \rightarrow \mathbb{R}^{H}, \\
    f_{\theta}(s,h,z)
    &=
    \mathrm{MLP}_{\theta}\big([\,s;\,h;\,z\,]\big).
\end{aligned}
\label{eq:app_vector_field}
\end{equation}
Here, $s$ is the normalized time coordinate, $h\in\mathbb{R}^{H}$ is
the latent state, and $z\in\mathbb{R}^{d_z}$ is an external control
input. In our configuration, $d_z=1$ for each independently processed
channel. The parameter set $\theta$ is shared across all samples and
across both decoder regimes.

\paragraph{Control path.}
The control is a time-dependent function
\begin{equation}
    z:
    [t_T,t_T+\Delta t]\rightarrow\mathbb{R}^{d_z}.
\end{equation}
At normalized integration coordinate $s$, the decoder evaluates the
control at the corresponding physical timestamp
$t(s)=t_T+s\Delta t$. The normalized time feature supplied to
$f_{\theta}$ and the physical-time argument supplied to $z(\cdot)$ are
therefore kept distinct.

\subsection{Unified Terminal Dynamics}
\label{app:cde_ode_unified}

Both decoder regimes are instances of the same initial-value problem.
Using the normalized variable $s$ and the Jacobian
$\mathrm{d}t/\mathrm{d}s=\Delta t$, the decoder solves
\begin{equation}
\begin{aligned}
    \frac{\mathrm{d}h}{\mathrm{d}s}(s)
    &=
    \Delta t\,
    f_{\theta}\!\left(
        s,
        h(s),
        z\!\left(t_T+s\Delta t\right)
    \right), \\
    h(0) &= h_T.
\end{aligned}
\label{eq:app_unified_dynamics}
\end{equation}
The endpoint state is
\begin{equation}
\begin{aligned}
    h(1)
    =
    h_T
    +
    \int_{0}^{1}
    \Delta t\,
    f_{\theta}\!\left(
        s,
        h(s),
        z\!\left(t_T+s\Delta t\right)
    \right)
    \mathrm{d}s.
\end{aligned}
\label{eq:app_unified_integral_s}
\end{equation}

For clarity, define the corresponding physical-time state
\begin{equation}
    \bar h(t)
    :=
    h\!\left(\frac{t-t_T}{\Delta t}\right).
    \label{eq:app_physical_state}
\end{equation}
Undoing the reparameterization in
Eq.~\eqref{eq:app_reparameterization} gives
\begin{equation}
\begin{aligned}
    \bar h(t_{T+1})
    =
    h_T
    +
    \int_{t_T}^{t_{T+1}}
    f_{\theta}\!\left(
        \frac{t-t_T}{\Delta t},
        \bar h(t),
        z(t)
    \right)
    \mathrm{d}t,
\end{aligned}
\label{eq:app_unified_integral_t}
\end{equation}
for non-degenerate intervals. Equations~\eqref{eq:app_unified_integral_s}
and~\eqref{eq:app_unified_integral_t} describe the same trajectory in
normalized and physical-time coordinates, respectively. The normalized
form in Eq.~\eqref{eq:app_unified_dynamics} is the one solved
numerically.

\subsection{The Two Decoder Regimes}
\label{app:cde_ode_regimes}

The forecasting and imputation regimes differ only in the control path
used in Eq.~\eqref{eq:app_unified_dynamics}.

\begin{definition}[Observation-guided regime]
\label{def:app_observation_guided}
For causal imputation, the available observations preceding the query
are interpolated using a natural cubic spline $S(\cdot)$, and the control
is defined as
\begin{equation}
    z(t)=S(t).
    \label{eq:app_spline_control}
\end{equation}
The latent dynamics are therefore conditioned on the observable causal
prefix through the external control input.
\end{definition}

\begin{definition}[Empty-control forecasting regime]
\label{def:app_empty_control}
For future forecasting, no observation-derived control is available
beyond the terminal context timestamp. The implementation represents
this absence of control using
\begin{equation}
    z(t)\equiv\mathbf{0}
    \in\mathbb{R}^{d_z}.
    \label{eq:app_zero_control}
\end{equation}
\end{definition}

The zero vector in Eq.~\eqref{eq:app_zero_control} is a valid input to
the same vector field in Eq.~\eqref{eq:app_vector_field}. Therefore,
switching from observation-guided imputation to forecasting does not
alter the decoder architecture or introduce a second parameter set.

\subsection{Exact Reduction}
\label{app:cde_ode_reduction}

We next show that the empty-control regime reduces exactly to a
time-augmented Neural ODE.

\paragraph{Input layer under zero control}
Let the first affine layer of $\mathrm{MLP}_{\theta}$ have weight
$W^{(1)}$ and bias $b^{(1)}$. Partition its columns according to the
three input blocks $[\,s;\,h;\,z\,]$:
\begin{equation}
    W^{(1)}
    =
    \left[
        W^{(1)}_{s}\;\;
        W^{(1)}_{h}\;\;
        W^{(1)}_{z}
    \right].
    \label{eq:app_weight_partition}
\end{equation}
The corresponding pre-activation is
\begin{equation}
\begin{aligned}
    a^{(1)}(s,h,z)
    =
    W^{(1)}_{s}s
    +
    W^{(1)}_{h}h
    +
    W^{(1)}_{z}z
    +
    b^{(1)}.
\end{aligned}
\label{eq:app_first_activation}
\end{equation}
When $z=\mathbf{0}$,
\begin{equation}
\begin{aligned}
    a^{(1)}(s,h,\mathbf{0})
    =
    W^{(1)}_{s}s
    +
    W^{(1)}_{h}h
    +
    b^{(1)},
\end{aligned}
\label{eq:app_first_activation_zero}
\end{equation}
because $W^{(1)}_{z}\mathbf{0}=\mathbf{0}$. Hence, under zero control,
the first-layer output and all subsequent activations depend only on
$(s,h)$ and the shared parameters $\theta$.

\paragraph{Reduced vector field}
Define
\begin{equation}
    \widetilde f_{\theta}(s,h)
    :=
    f_{\theta}(s,h,\mathbf{0}).
    \label{eq:app_reduced_field}
\end{equation}
Substituting Eq.~\eqref{eq:app_zero_control} into
Eq.~\eqref{eq:app_unified_dynamics} yields
\begin{equation}
\begin{aligned}
    \frac{\mathrm{d}h}{\mathrm{d}s}(s)
    &=
    \Delta t\,
    \widetilde f_{\theta}\big(s,h(s)\big), \\
    h(0)&=h_T.
\end{aligned}
\label{eq:app_ode_dynamics}
\end{equation}
Its endpoint solution is
\begin{equation}
\begin{aligned}
    h(1)
    =
    h_T
    +
    \int_{0}^{1}
    \Delta t\,
    \widetilde f_{\theta}\big(s,h(s)\big)
    \mathrm{d}s.
\end{aligned}
\label{eq:app_ode_integral}
\end{equation}

Equation~\eqref{eq:app_ode_dynamics} is a time-augmented Neural ODE:
its vector field depends only on the normalized time coordinate and the
latent state. It uses the same parameter set $\theta$ as
Eq.~\eqref{eq:app_unified_dynamics}; the control-related column block
$W^{(1)}_{z}$ remains part of the shared network but contributes zero
under Eq.~\eqref{eq:app_zero_control}.

The reduction follows directly and pointwise:
\begin{equation}
\begin{aligned}
    f_{\theta}
    \left(
        s,
        h(s),
        z(t_T+s\Delta t)
    \right)
    &=
    f_{\theta}\big(s,h(s),\mathbf{0}\big) \\
    &=
    \widetilde f_{\theta}\big(s,h(s)\big).
\end{aligned}
\label{eq:app_pointwise_reduction}
\end{equation}
No parameter is added, removed, or updated when switching between the
two regimes. Therefore, the reduction is an exact algebraic identity,
not an approximation induced by the numerical solver.

\paragraph{Well-posedness}
Suppose that $f_{\theta}$ is continuous in $(s,z)$ and uniformly
Lipschitz continuous in $h$ over the compact integration interval. The
natural cubic spline control is continuous, while the empty control is
constant. Under these conditions, the Picard--Lindel\"of theorem ensures
that both Eq.~\eqref{eq:app_unified_dynamics} and its reduction in
Eq.~\eqref{eq:app_ode_dynamics} admit unique solutions. In particular,
fixing $z\equiv\mathbf{0}$ preserves the Lipschitz property in $h$ and
requires no additional well-posedness assumptions. An MLP with finite
weights and Lipschitz activations satisfies the required Lipschitz
condition.

\paragraph{Globally disabled control channel}
If the decoder is instantiated with $d_z=0$, the control block is omitted
from the concatenated input, and its vector field takes the form
\begin{equation}
    f_{\theta}(s,h)
    =
    \mathrm{MLP}_{\theta}\big([\,s;\,h\,]\big).
\end{equation}
Functionally, this is the same reduced form as
Eq.~\eqref{eq:app_reduced_field}, with the zero-contributing control
columns omitted at construction time. Thus, both a runtime empty
control path and a globally disabled control channel result in
time-augmented Neural ODE dynamics.

\paragraph{Terminology}
The formulation in Eq.~\eqref{eq:app_vector_field} injects the control
value $z(t)$ as an input to the vector field:
\begin{equation}
    \frac{\mathrm{d}h}{\mathrm{d}t}
    =
    f_{\theta}\big(\cdot,h(t),z(t)\big).
\end{equation}
This is a control-as-input neural differential equation. It is related
to, but mathematically distinct from, the commonly used bilinear Neural
CDE formulation
\begin{equation}
    \mathrm{d}h(t)
    =
    g_{\theta}\big(h(t)\big)\,\mathrm{d}X(t),
    \label{eq:app_bilinear_cde}
\end{equation}
in which the vector field acts on increments of the control path. When
$X$ is differentiable, the latter can be written as
\begin{equation}
    \frac{\mathrm{d}h}{\mathrm{d}t}
    =
    g_{\theta}\big(h(t)\big)\dot X(t).
\end{equation}
SOTER instead supplies the control value jointly with time and state to
a single MLP. Throughout this work, Neural CDE-style dynamics refer to this control-as-input controlled neural differential equation. The
zero-control reduction established above applies specifically to this
formulation.

\paragraph{Task interpretation}
The reduction is consistent with the semantics of the two downstream
tasks. In future forecasting, no observation exists beyond the context
boundary from which to construct a control path; setting
$z\equiv\mathbf{0}$ therefore yields autonomous extrapolation from the
terminal state $h_T$. In causal imputation, the observations available
before the query define the spline control $z=S(\cdot)$, which guides
the same latent dynamics toward the missing timestamp.

\paragraph{Numerical solver}
Both regimes are integrated over $s\in[0,1]$ using the same adaptive Dormand--Prince 5(4) solver (dopri5), with absolute and
relative tolerances of $10^{-5}$. The selection of the solver and tolerance is detailed in Appendix~\ref{app:sensitivity}. The algebraic reduction is independent
of the numerical solver and its tolerance; these choices affect only
the numerical approximation of
Eqs.~\eqref{eq:app_unified_integral_s} and~\eqref{eq:app_ode_integral}.

\subsection{Summary}
\label{app:cde_ode_summary}

The forecasting decoder in Eq.~\eqref{eq:app_ode_dynamics} is obtained
by evaluating the observation-guided decoder in
Eq.~\eqref{eq:app_unified_dynamics} at the zero control path. The two
regimes share the same network structure and parameter set $\theta$.
Consequently, a single frozen continuous-time decoder supports both
observation-guided causal imputation and autonomous future forecasting
without switching models or introducing task-specific decoder
parameters.

\section{System Overhead Analysis}
\label{app:system_overhead}

To characterize the practical inference overhead, we measure the average
single-sample latency of representative models under the same hardware
and software environment and report their parameter counts using the
same convention as in the main paper. TimeMoE-Large and TimeMoE-Base
require $3.72$~s and $3.92$~s per sample, with $453.2$M and $113.3$M
parameters, respectively. MIRA-Large contains $339.9$M parameters and
incurs the highest latency of $8.97$~s, whereas Aurora achieves the
lowest latency of $0.138$~s with $260.8$M parameters. SOTER requires
$6.09$~s per sample while activating only $20.29$M parameters. It is
therefore slower than Aurora and the two TimeMoE variants, but faster
than MIRA-Large and substantially more compact than all compared models.
These results also show that wall-clock latency is not determined by
parameter count alone: SOTER's adaptive continuous-time integration
introduces non-negligible numerical overhead, while architectural
structure and implementation characteristics substantially influence
runtime. The comparison should thus be interpreted as a practical
accuracy--compactness--latency trade-off rather than a universal ranking
of inference efficiency.

\section{Complete Per-Window Zero-Shot Forecasting Results}
\label{app:forecasting_results}

This section reports the per-window zero-shot forecasting measurements
underlying Table~\ref{tab:main_results}. Unlike the main table, which
reports the mean and sample standard deviation across the supported
input/output windows after rescaling all error values by $10^{2}$ for
readability, Tables~\ref{tab:e-win24}--\ref{tab:e-win64} report the
unscaled RMSE and MAE values in the normalized space for each individual
window. The aggregate statistics in Table~\ref{tab:main_results} are
computed from the underlying full-precision measurements before
rounding; consequently, recomputing them from the four-decimal values
reported below may produce a difference in the final displayed digit.
Timer supports only the two longest input/output windows, $96/48$ and
$128/64$.

\begin{table*}[t!]
    \caption{Per-window zero-shot forecasting results for the $48/24$
    input/output setting. Unscaled RMSE and MAE values are reported in
    the normalized space. Timer does not support this setting and is
    marked ``/''. Bold and underlined values denote the best and
    second-best results in each column, respectively.}
    \label{tab:e-win24}
    \centering
    \footnotesize
    \renewcommand{\arraystretch}{0.9}
    \setlength{\tabcolsep}{3.5pt}
    \resizebox{\textwidth}{!}{%
    \begin{tabular}{lcccccccccccc}
        \toprule
        \multirow{2}{*}{Method} & \multicolumn{2}{c}{\shortstack{HeartRate\\(SC)}} & \multicolumn{2}{c}{\shortstack{MIT-BIH\\(MC)}} & \multicolumn{2}{c}{\shortstack{ScientiSST\\MOVE (MC)}} & \multicolumn{2}{c}{\shortstack{IEEE PPG\\(MC)}} & \multicolumn{2}{c}{\shortstack{MMASH\\(MC)}} & \multicolumn{2}{c}{\shortstack{WDD\\(MC)}} \\
        \cmidrule(lr){2-3}\cmidrule(lr){4-5}\cmidrule(lr){6-7}\cmidrule(lr){8-9}\cmidrule(lr){10-11}\cmidrule(lr){12-13}
        & RMSE & MAE & RMSE & MAE & RMSE & MAE & RMSE & MAE & RMSE & MAE & RMSE & MAE \\
        \midrule
        Timer & / & / & / & / & / & / & / & / & / & / & / & / \\
        TTM & 0.0397 & 0.0214 & 0.0398 & 0.0192 & 0.0948 & 0.0715 & 0.0934 & 0.0629 & 0.0106 & 0.0079 & 0.0207 & 0.0167 \\
        Lag-Llama & 0.1967 & 0.1554 & 0.4906 & 0.4876 & 0.5134 & 0.4726 & 0.5187 & 0.5082 & 0.1849 & 0.1769 & 0.4348 & 0.4145 \\
        TiRex & 0.0367 & 0.0171 & 0.0448 & 0.0208 & 0.0634 & 0.0467 & 0.0919 & 0.0536 & 0.0010 & \underline{0.0003} & 0.0106 & 0.0078 \\
        MOMENT-Large & 0.0620 & 0.0366 & 0.0419 & 0.0220 & 0.1147 & 0.0927 & 0.1022 & 0.0618 & 0.0036 & 0.0015 & 0.0114 & \underline{0.0060} \\
        MOMENT-Base & 0.0685 & 0.0429 & 0.0470 & 0.0247 & 0.1157 & 0.0920 & 0.1024 & 0.0621 & 0.0032 & 0.0016 & 0.0113 & 0.0063 \\
        MOMENT-Small & 0.0651 & 0.0392 & 0.0426 & 0.0223 & 0.1130 & 0.0902 & 0.1049 & 0.0633 & 0.0044 & 0.0017 & 0.0109 & \textbf{0.0058} \\
        MoiraiMoE-Base & 0.0583 & 0.0246 & 0.0512 & 0.0228 & 0.1397 & 0.0963 & 0.1251 & 0.0696 & 0.0022 & 0.0009 & 0.0114 & 0.0101 \\
        MoiraiMoE-Small & 0.0619 & 0.0280 & 0.0509 & 0.0227 & 0.1430 & 0.0992 & 0.1296 & 0.0713 & 0.0026 & 0.0010 & 0.0112 & 0.0097 \\
        Moirai2.0-Small & 0.0527 & 0.0214 & 0.0479 & 0.0217 & 0.0877 & 0.0432 & 0.0947 & 0.0548 & \underline{0.0007} & \underline{0.0003} & 0.0096 & 0.0084 \\
        TimesFM 1.0 & 0.0472 & 0.0231 & 0.0441 & 0.0218 & 0.1165 & 0.0853 & 0.0950 & 0.0538 & 0.0027 & 0.0008 & 0.0090 & 0.0066 \\
        TimesFM 2.0 & 0.0545 & 0.0256 & 0.0464 & 0.0211 & 0.1170 & 0.0832 & 0.1054 & 0.0585 & 0.0010 & \underline{0.0003} & \underline{0.0089} & 0.0065 \\
        TimesFM 2.5 & 0.0521 & 0.0218 & 0.0429 & 0.0201 & 0.1182 & 0.0833 & 0.0992 & 0.0559 & \textbf{0.0003} & \textbf{0.0001} & 0.0093 & 0.0072 \\
        TimeMoE-Large & 0.0266 & 0.0163 & \underline{0.0197} & \underline{0.0099} & \underline{0.0319} & \underline{0.0203} & \underline{0.0173} & \textbf{0.0108} & 0.0339 & 0.0288 & 0.0167 & 0.0125 \\
        TimeMoE-Base & \underline{0.0234} & \underline{0.0162} & 0.0206 & 0.0123 & 0.0332 & 0.0213 & \textbf{0.0172} & \underline{0.0112} & 0.0358 & 0.0287 & 0.0204 & 0.0162 \\
        Chronos2-Base & 0.0523 & 0.0236 & 0.0632 & 0.0238 & 0.0801 & 0.0410 & 0.1090 & 0.0616 & 0.0012 & 0.0005 & 0.0099 & 0.0084 \\
        Chronos2-Small & 0.0549 & 0.0252 & 0.0519 & 0.0224 & 0.1156 & 0.0798 & 0.1034 & 0.0597 & 0.0015 & 0.0006 & 0.0098 & 0.0083 \\
        Sundial-Base & 0.0678 & 0.0313 & 0.0409 & 0.0198 & 0.1122 & 0.0822 & 0.0907 & 0.0525 & 0.0156 & 0.0124 & 0.0151 & 0.0109 \\
        Aurora & 0.0853 & 0.0456 & 0.0509 & 0.0274 & 0.1604 & 0.1262 & 0.1083 & 0.0671 & 0.0197 & 0.0085 & 0.0107 & 0.0062 \\
        MIRA-Large & 0.0478 & 0.0343 & 0.0234 & 0.0167 & 0.0494 & 0.0387 & 0.0266 & 0.0204 & 0.0653 & 0.0581 & 0.0334 & 0.0295 \\
        \midrule
        \rowcolor[HTML]{E5E5E5}
        SOTER & \textbf{0.0150} & \textbf{0.0106} & \textbf{0.0157} & \textbf{0.0075} & \textbf{0.0244} & \textbf{0.0121} & 0.0246 & \underline{0.0112} & 0.0163 & 0.0104 & \textbf{0.0086} & 0.0072 \\
        \bottomrule
    \end{tabular}%
    }
\end{table*}

\begin{table*}[t!]
    \caption{Per-window zero-shot forecasting results for the $72/36$
    input/output setting. Unscaled RMSE and MAE values are reported in
    the normalized space. Timer does not support this setting and is
    marked ``/''. Bold and underlined values denote the best and
    second-best results in each column, respectively.}
    \label{tab:e-win36}
    \centering
    \footnotesize
    \renewcommand{\arraystretch}{0.9}
    \setlength{\tabcolsep}{3.5pt}
    \resizebox{\textwidth}{!}{%
    \begin{tabular}{lcccccccccccc}
        \toprule
        \multirow{2}{*}{Method} & \multicolumn{2}{c}{\shortstack{HeartRate\\(SC)}} & \multicolumn{2}{c}{\shortstack{MIT-BIH\\(MC)}} & \multicolumn{2}{c}{\shortstack{ScientiSST\\MOVE (MC)}} & \multicolumn{2}{c}{\shortstack{IEEE PPG\\(MC)}} & \multicolumn{2}{c}{\shortstack{MMASH\\(MC)}} & \multicolumn{2}{c}{\shortstack{WDD\\(MC)}} \\
        \cmidrule(lr){2-3}\cmidrule(lr){4-5}\cmidrule(lr){6-7}\cmidrule(lr){8-9}\cmidrule(lr){10-11}\cmidrule(lr){12-13}
        & RMSE & MAE & RMSE & MAE & RMSE & MAE & RMSE & MAE & RMSE & MAE & RMSE & MAE \\
        \midrule
        Timer & / & / & / & / & / & / & / & / & / & / & / & / \\
        TTM & 0.0485 & 0.0271 & 0.0418 & 0.0205 & 0.0959 & 0.0739 & 0.0929 & 0.0624 & 0.0120 & 0.0074 & 0.0175 & 0.0124 \\
        Lag-Llama & 0.1989 & 0.1567 & 0.4886 & 0.4857 & 0.5063 & 0.4649 & 0.5165 & 0.5059 & 0.1837 & 0.1760 & 0.4385 & 0.4146 \\
        TiRex & 0.0526 & 0.0272 & 0.0407 & 0.0203 & 0.0612 & 0.0426 & 0.0927 & 0.0567 & 0.0025 & 0.0009 & 0.0099 & 0.0084 \\
        MOMENT-Large & 0.0589 & 0.0311 & 0.0440 & 0.0226 & 0.0894 & 0.0734 & 0.0978 & 0.0618 & 0.0052 & 0.0024 & 0.0107 & \underline{0.0060} \\
        MOMENT-Base & 0.0685 & 0.0361 & 0.0521 & 0.0258 & 0.0930 & 0.0729 & 0.1018 & 0.0638 & 0.0048 & 0.0022 & 0.0108 & 0.0063 \\
        MOMENT-Small & 0.0624 & 0.0311 & 0.0449 & 0.0227 & 0.0881 & 0.0708 & 0.1036 & 0.0649 & 0.0049 & 0.0022 & 0.0108 & 0.0062 \\
        MoiraiMoE-Base & 0.0712 & 0.0342 & 0.0622 & 0.0279 & 0.1352 & 0.0971 & 0.1207 & 0.0734 & 0.0054 & 0.0022 & 0.0125 & 0.0107 \\
        MoiraiMoE-Small & 0.0725 & 0.0345 & 0.0563 & 0.0256 & 0.1349 & 0.0967 & 0.1213 & 0.0728 & 0.0060 & 0.0025 & 0.0118 & 0.0099 \\
        Moirai2.0-Small & 0.0506 & 0.0307 & 0.0506 & 0.0234 & 0.0789 & 0.0435 & 0.1021 & 0.0617 & 0.0070 & 0.0026 & 0.0098 & 0.0085 \\
        TimesFM 1.0 & 0.0708 & 0.0317 & 0.0426 & 0.0207 & 0.1154 & 0.0832 & 0.0820 & 0.0527 & 0.0035 & 0.0010 & 0.0095 & 0.0068 \\
        TimesFM 2.0 & 0.0708 & 0.0331 & 0.0469 & 0.0215 & 0.1154 & 0.0836 & 0.0908 & 0.0567 & 0.0023 & \underline{0.0007} & 0.0094 & 0.0068 \\
        TimesFM 2.5 & 0.0563 & 0.0260 & 0.0418 & 0.0201 & 0.1161 & 0.0849 & 0.0857 & 0.0535 & \textbf{0.0007} & \textbf{0.0003} & \underline{0.0092} & 0.0074 \\
        TimeMoE-Large & \underline{0.0154} & \underline{0.0108} & \underline{0.0179} & \underline{0.0096} & \underline{0.0315} & \underline{0.0199} & \textbf{0.0187} & \underline{0.0109} & 0.0305 & 0.0262 & 0.0157 & 0.0121 \\
        TimeMoE-Base & 0.0176 & 0.0125 & 0.0197 & 0.0124 & 0.0353 & 0.0225 & \underline{0.0188} & 0.0113 & 0.0310 & 0.0255 & 0.0206 & 0.0159 \\
        Chronos2-Base & 0.0464 & 0.0208 & 0.0495 & 0.0234 & 0.0747 & 0.0402 & 0.1018 & 0.0635 & \underline{0.0012} & \underline{0.0007} & 0.0109 & 0.0062 \\
        Chronos2-Small & 0.0611 & 0.0272 & 0.0507 & 0.0244 & 0.1175 & 0.0813 & 0.1041 & 0.0647 & 0.0019 & 0.0008 & 0.0108 & 0.0062 \\
        Sundial-Base & 0.0637 & 0.0296 & 0.0423 & 0.0201 & 0.1061 & 0.0793 & 0.0841 & 0.0525 & 0.0119 & 0.0107 & 0.0149 & 0.0102 \\
        Aurora & 0.0622 & 0.0362 & 0.0541 & 0.0254 & 0.1282 & 0.0991 & 0.0978 & 0.0625 & 0.0182 & 0.0074 & 0.0100 & \textbf{0.0056} \\
        MIRA-Large & 0.0285 & 0.0181 & 0.0228 & 0.0158 & 0.0483 & 0.0376 & 0.0253 & 0.0202 & 0.0628 & 0.0559 & 0.0334 & 0.0301 \\
        \midrule
        \rowcolor[HTML]{E5E5E5}
        SOTER & \textbf{0.0119} & \textbf{0.0080} & \textbf{0.0136} & \textbf{0.0064} & \textbf{0.0223} & \textbf{0.0119} & 0.0217 & \textbf{0.0104} & 0.0127 & 0.0098 & \textbf{0.0075} & 0.0062 \\
        \bottomrule
    \end{tabular}%
    }
\end{table*}

\begin{table*}[t!]
    \caption{Per-window zero-shot forecasting results for the $96/48$
    input/output setting. Unscaled RMSE and MAE values are reported in
    the normalized space. Bold and underlined values denote the best and
    second-best results in each column, respectively.}
    \label{tab:e-win48}
    \centering
    \footnotesize
    \renewcommand{\arraystretch}{0.9}
    \setlength{\tabcolsep}{3.5pt}
    \resizebox{\textwidth}{!}{%
    \begin{tabular}{lcccccccccccc}
        \toprule
        \multirow{2}{*}{Method} & \multicolumn{2}{c}{\shortstack{HeartRate\\(SC)}} & \multicolumn{2}{c}{\shortstack{MIT-BIH\\(MC)}} & \multicolumn{2}{c}{\shortstack{ScientiSST\\MOVE (MC)}} & \multicolumn{2}{c}{\shortstack{IEEE PPG\\(MC)}} & \multicolumn{2}{c}{\shortstack{MMASH\\(MC)}} & \multicolumn{2}{c}{\shortstack{WDD\\(MC)}} \\
        \cmidrule(lr){2-3}\cmidrule(lr){4-5}\cmidrule(lr){6-7}\cmidrule(lr){8-9}\cmidrule(lr){10-11}\cmidrule(lr){12-13}
        & RMSE & MAE & RMSE & MAE & RMSE & MAE & RMSE & MAE & RMSE & MAE & RMSE & MAE \\
        \midrule
        Timer & 0.0801 & 0.0448 & 0.0629 & 0.0365 & 0.1559 & 0.1193 & 0.1254 & 0.0843 & 0.0272 & 0.0209 & 0.0325 & 0.0263 \\
        TTM & 0.0550 & 0.0309 & 0.0420 & 0.0211 & 0.0966 & 0.0756 & 0.0895 & 0.0582 & 0.0163 & 0.0102 & 0.0216 & 0.0166 \\
        Lag-Llama & 0.1985 & 0.1551 & 0.4895 & 0.4865 & 0.5133 & 0.4727 & 0.5176 & 0.5070 & 0.1829 & 0.1754 & 0.4385 & 0.4146 \\
        TiRex & 0.0674 & 0.0299 & 0.0432 & 0.0212 & 0.0563 & 0.0370 & 0.0872 & 0.0535 & 0.0048 & 0.0024 & \underline{0.0089} & 0.0082 \\
        MOMENT-Large & 0.0685 & 0.0334 & 0.0418 & 0.0215 & 0.0918 & 0.0757 & 0.1106 & 0.0697 & 0.0082 & 0.0039 & 0.0129 & 0.0067 \\
        MOMENT-Base & 0.0739 & 0.0386 & 0.0478 & 0.0258 & 0.0908 & 0.0718 & 0.1146 & 0.0716 & 0.0083 & 0.0039 & 0.0139 & 0.0075 \\
        MOMENT-Small & 0.0706 & 0.0349 & 0.0441 & 0.0234 & 0.0885 & 0.0704 & 0.1150 & 0.0718 & 0.0082 & 0.0038 & 0.0129 & 0.0066 \\
        MoiraiMoE-Base & 0.0762 & 0.0374 & 0.0554 & 0.0258 & 0.1285 & 0.0922 & 0.1095 & 0.0687 & 0.0101 & 0.0038 & 0.0130 & 0.0103 \\
        MoiraiMoE-Small & 0.0801 & 0.0406 & 0.0548 & 0.0253 & 0.1329 & 0.0963 & 0.1143 & 0.0708 & 0.0092 & 0.0035 & 0.0132 & 0.0109 \\
        Moirai2.0-Small & 0.0572 & 0.0283 & 0.0456 & 0.0219 & 0.0798 & 0.0418 & 0.0842 & 0.0532 & \underline{0.0019} & \underline{0.0007} & 0.0112 & 0.0092 \\
        TimesFM 1.0 & 0.0644 & 0.0296 & 0.0446 & 0.0219 & 0.1034 & 0.0771 & 0.0818 & 0.0519 & 0.0032 & 0.0011 & 0.0108 & 0.0071 \\
        TimesFM 2.0 & 0.0710 & 0.0338 & 0.0447 & 0.0216 & 0.1079 & 0.0805 & 0.0868 & 0.0558 & 0.0026 & 0.0010 & 0.0102 & 0.0069 \\
        TimesFM 2.5 & 0.0627 & 0.0278 & 0.0431 & 0.0206 & 0.1071 & 0.0781 & 0.0821 & 0.0517 & \textbf{0.0008} & \textbf{0.0003} & 0.0110 & 0.0084 \\
        TimeMoE-Large & \textbf{0.0129} & \underline{0.0097} & \underline{0.0183} & \underline{0.0102} & \underline{0.0309} & \underline{0.0193} & \textbf{0.0176} & \underline{0.0106} & 0.0295 & 0.0252 & 0.0153 & 0.0119 \\
        TimeMoE-Base & 0.0170 & 0.0117 & 0.0218 & 0.0144 & 0.0359 & 0.0226 & \underline{0.0180} & 0.0114 & 0.0341 & 0.0266 & 0.0230 & 0.0174 \\
        Chronos2-Base & 0.0702 & 0.0303 & 0.0449 & 0.0209 & 0.0699 & 0.0387 & 0.0852 & 0.0534 & 0.0024 & 0.0009 & 0.0110 & \textbf{0.0053} \\
        Chronos2-Small & 0.0546 & 0.0319 & 0.0445 & 0.0212 & 0.0999 & 0.0689 & 0.0854 & 0.0536 & 0.0031 & 0.0012 & 0.0112 & \underline{0.0055} \\
        Sundial-Base & 0.0459 & 0.0244 & 0.0430 & 0.0206 & 0.1117 & 0.0847 & 0.0794 & 0.0512 & 0.0101 & 0.0092 & 0.0130 & 0.0091 \\
        Aurora & 0.0731 & 0.0398 & 0.0494 & 0.0258 & 0.1424 & 0.1097 & 0.1014 & 0.0650 & 0.0179 & 0.0078 & 0.0121 & 0.0063 \\
        MIRA-Large & 0.0164 & 0.0116 & 0.0214 & 0.0139 & 0.0471 & 0.0368 & 0.0252 & 0.0204 & 0.0615 & 0.0537 & 0.0340 & 0.0302 \\
        \midrule
        \rowcolor[HTML]{E5E5E5}
        SOTER & \underline{0.0145} & \textbf{0.0093} & \textbf{0.0114} & \textbf{0.0056} & \textbf{0.0219} & \textbf{0.0117} & 0.0207 & \textbf{0.0099} & 0.0112 & 0.0082 & \textbf{0.0067} & \underline{0.0055} \\
        \bottomrule
    \end{tabular}%
    }
\end{table*}

\begin{table*}[t!]
    \caption{Per-window zero-shot forecasting results for the $128/64$
    input/output setting. Unscaled RMSE and MAE values are reported in
    the normalized space. Bold and underlined values denote the best and
    second-best results in each column, respectively.}
    \label{tab:e-win64}
    \centering
    \footnotesize
    \renewcommand{\arraystretch}{0.9}
    \setlength{\tabcolsep}{3.5pt}
    \resizebox{\textwidth}{!}{%
    \begin{tabular}{lcccccccccccc}
        \toprule
        \multirow{2}{*}{Method} & \multicolumn{2}{c}{\shortstack{HeartRate\\(SC)}} & \multicolumn{2}{c}{\shortstack{MIT-BIH\\(MC)}} & \multicolumn{2}{c}{\shortstack{ScientiSST\\MOVE (MC)}} & \multicolumn{2}{c}{\shortstack{IEEE PPG\\(MC)}} & \multicolumn{2}{c}{\shortstack{MMASH\\(MC)}} & \multicolumn{2}{c}{\shortstack{WDD\\(MC)}} \\
        \cmidrule(lr){2-3}\cmidrule(lr){4-5}\cmidrule(lr){6-7}\cmidrule(lr){8-9}\cmidrule(lr){10-11}\cmidrule(lr){12-13}
        & RMSE & MAE & RMSE & MAE & RMSE & MAE & RMSE & MAE & RMSE & MAE & RMSE & MAE \\
        \midrule
        Timer & 0.0979 & 0.0521 & 0.0593 & 0.0363 & 0.1560 & 0.1192 & 0.1330 & 0.0898 & 0.0311 & 0.0228 & 0.0339 & 0.0276 \\
        TTM & 0.0589 & 0.0332 & 0.0419 & 0.0216 & 0.0967 & 0.0762 & 0.0900 & 0.0589 & 0.0204 & 0.0121 & 0.0232 & 0.0181 \\
        Lag-Llama & 0.2041 & 0.1580 & 0.4887 & 0.4857 & 0.5079 & 0.4667 & 0.5173 & 0.5063 & 0.1832 & 0.1755 & 0.4382 & 0.4143 \\
        TiRex & 0.0521 & 0.0256 & 0.0374 & 0.0184 & 0.0517 & 0.0358 & 0.0853 & 0.0502 & 0.0066 & 0.0030 & \underline{0.0097} & 0.0063 \\
        MOMENT-Large & 0.0636 & 0.0333 & 0.0417 & 0.0221 & 0.0937 & 0.0741 & 0.1133 & 0.0710 & 0.0105 & 0.0047 & 0.0161 & 0.0093 \\
        MOMENT-Base & 0.0668 & 0.0352 & 0.0451 & 0.0246 & 0.0946 & 0.0738 & 0.1177 & 0.0730 & 0.0096 & 0.0043 & 0.0170 & 0.0096 \\
        MOMENT-Small & 0.0657 & 0.0335 & 0.0436 & 0.0234 & 0.0892 & 0.0705 & 0.1189 & 0.0738 & 0.0107 & 0.0049 & 0.0166 & 0.0091 \\
        MoiraiMoE-Base & 0.0744 & 0.0411 & 0.0501 & 0.0237 & 0.1237 & 0.0844 & 0.1118 & 0.0706 & 0.0140 & 0.0052 & 0.0138 & 0.0109 \\
        MoiraiMoE-Small & 0.0775 & 0.0428 & 0.0479 & 0.0227 & 0.1265 & 0.0919 & 0.1155 & 0.0716 & 0.0129 & 0.0049 & 0.0139 & 0.0112 \\
        Moirai2.0-Small & 0.0526 & 0.0254 & 0.0395 & 0.0194 & 0.0785 & 0.0407 & 0.0867 & 0.0548 & \underline{0.0031} & \underline{0.0012} & 0.0123 & 0.0097 \\
        TimesFM 1.0 & 0.0590 & 0.0270 & 0.0439 & 0.0212 & 0.0944 & 0.0719 & 0.0854 & 0.0538 & 0.0039 & 0.0014 & 0.0121 & 0.0083 \\
        TimesFM 2.0 & 0.0623 & 0.0301 & 0.0389 & 0.0192 & 0.1004 & 0.0762 & 0.0868 & 0.0559 & 0.0032 & \underline{0.0012} & 0.0117 & 0.0072 \\
        TimesFM 2.5 & 0.0508 & 0.0235 & 0.0364 & 0.0181 & 0.0934 & 0.0681 & 0.0845 & 0.0530 & \textbf{0.0013} & \textbf{0.0005} & 0.0125 & 0.0093 \\
        TimeMoE-Large & \textbf{0.0123} & \textbf{0.0097} & \underline{0.0171} & \underline{0.0101} & \underline{0.0299} & \underline{0.0187} & \textbf{0.0170} & \underline{0.0107} & 0.0283 & 0.0241 & 0.0152 & 0.0117 \\
        TimeMoE-Base & 0.0160 & 0.0116 & 0.0237 & 0.0167 & 0.0360 & 0.0225 & \underline{0.0176} & 0.0117 & 0.0314 & 0.0252 & 0.0225 & 0.0172 \\
        Chronos2-Base & 0.0570 & 0.0250 & 0.0345 & 0.0164 & 0.0636 & 0.0354 & 0.0869 & 0.0531 & 0.0046 & 0.0017 & 0.0117 & \underline{0.0052} \\
        Chronos2-Small & 0.0562 & 0.0251 & 0.0345 & 0.0164 & 0.0851 & 0.0570 & 0.0858 & 0.0529 & 0.0052 & 0.0019 & 0.0121 & 0.0054 \\
        Sundial-Base & 0.0525 & 0.0254 & 0.0386 & 0.0194 & 0.0999 & 0.0526 & 0.0819 & 0.0526 & 0.0080 & 0.0065 & 0.0122 & 0.0077 \\
        Aurora & 0.0647 & 0.0370 & 0.0495 & 0.0256 & 0.1206 & 0.0926 & 0.1044 & 0.0662 & 0.0237 & 0.0089 & 0.0139 & 0.0067 \\
        MIRA-Large & \underline{0.0136} & \underline{0.0100} & 0.0203 & 0.0153 & 0.0456 & 0.0359 & 0.0258 & 0.0205 & 0.0596 & 0.0519 & 0.0335 & 0.0300 \\
        \midrule
        \rowcolor[HTML]{E5E5E5}
        SOTER & 0.0152 & 0.0101 & \textbf{0.0109} & \textbf{0.0054} & \textbf{0.0217} & \textbf{0.0112} & 0.0204 & \textbf{0.0103} & 0.0103 & 0.0071 & \textbf{0.0064} & \textbf{0.0051} \\
        \bottomrule
    \end{tabular}%
    }
\end{table*}

\section{Continuous-Time Imputation across Missing Rates}
\label{app:imputation_results}

Figure~\ref{fig:imputation} in the main paper visualizes the mean
continuous-time imputation errors over three repeated runs at missing
rates of $25\%$, $50\%$, and $75\%$. For visual clarity, the figure
plots only the mean value of each model and configuration. In contrast,
Tables~\ref{tab:e-imp25}--\ref{tab:e-imp75} report the complete
numerical results, including the corresponding standard deviations and
the full set of evaluated baselines. All entries are reported as the
mean and standard deviation over three runs with different random seeds
in the normalized space.

At missing rates of $25\%$ and $50\%$, SOTER achieves the best result
on 10 of the 12 dataset--metric combinations and ranks second on the
remaining two, both on MMASH, where TiRex performs best. At a missing
rate of $75\%$, SOTER achieves the lowest error on all 12 combinations.
Several baselines also exhibit substantially greater degradation as the
missing rate increases. For example, on MIT-BIH, the RMSE of
Chronos2-Base increases from $0.0578$ at $25\%$ missingness to $0.4850$
at $75\%$, while that of Sundial-Base increases from $0.1063$ to
$0.4105$. Over the same range, SOTER increases from $0.0183$ to
$0.1502$. These results are consistent with the ability of SOTER's
continuous-time decoder to query a target timestamp directly from the
available causal prefix.

\begin{table*}[t!]
    \caption{Complete continuous-time imputation results at a $25\%$
    missing rate. Each entry reports the mean and standard deviation
    over three runs with different random seeds in the normalized space.
    The mean values are used to construct Figure~\ref{fig:imputation}.
    Bold and underlined means denote the best and second-best results in
    each column, respectively.}
    \label{tab:e-imp25}
    \centering
    \footnotesize
    \renewcommand{\arraystretch}{1.0}
    \setlength{\tabcolsep}{2.5pt}
    \resizebox{\textwidth}{!}{%
    \begin{tabular}{lcccccccccccc}
        \toprule
        \multirow{2}{*}{Method} & \multicolumn{2}{c}{\shortstack{HeartRate\\(SC)}} & \multicolumn{2}{c}{\shortstack{MIT-BIH\\(MC)}} & \multicolumn{2}{c}{\shortstack{ScientiSST\\MOVE (MC)}} & \multicolumn{2}{c}{\shortstack{IEEE PPG\\(MC)}} & \multicolumn{2}{c}{\shortstack{MMASH\\(MC)}} & \multicolumn{2}{c}{\shortstack{WDD\\(MC)}} \\
        \cmidrule(lr){2-3}\cmidrule(lr){4-5}\cmidrule(lr){6-7}\cmidrule(lr){8-9}\cmidrule(lr){10-11}\cmidrule(lr){12-13}
        & RMSE & MAE & RMSE & MAE & RMSE & MAE & RMSE & MAE & RMSE & MAE & RMSE & MAE \\
        \midrule
        TimeMoE-Large & \pmstd{0.1785}{0.0018} & \pmstd{0.1615}{0.0014} & \pmstd{0.1748}{0.0013} & \pmstd{0.1603}{0.0006} & \pmstd{0.1801}{0.0035} & \pmstd{0.1598}{0.0017} & \pmstd{0.1891}{0.0021} & \pmstd{0.1649}{0.0008} & \pmstd{0.1347}{0.0024} & \pmstd{0.1268}{0.0013} & \pmstd{0.1785}{0.0018} & \pmstd{0.1614}{0.0013} \\
        Moirai2.0-Small & \pmstd{0.0510}{0.0122} & \pmstd{0.0281}{0.0054} & \pmstd{0.1083}{0.0004} & \pmstd{0.0724}{0.0002} & \pmstd{0.0819}{0.0016} & \pmstd{0.0474}{0.0012} & \pmstd{0.1214}{0.0026} & \pmstd{0.0797}{0.0016} & \pmstd{0.0292}{0.0005} & \pmstd{0.0175}{0.0004} & \pmstd{0.0460}{0.0022} & \pmstd{0.0261}{0.0014} \\
        MOMENT-Large & \pmstd{0.0535}{0.0043} & \pmstd{0.0322}{0.0006} & \pmstd{0.1148}{0.0123} & \pmstd{0.1003}{0.0179} & \pmstd{0.1018}{0.0047} & \pmstd{0.0798}{0.0073} & \pmstd{0.1698}{0.0167} & \pmstd{0.1425}{0.0177} & \pmstd{0.0436}{0.0023} & \pmstd{0.0340}{0.0018} & \pmstd{0.1117}{0.0071} & \pmstd{0.0943}{0.0052} \\
        Chronos2-Base & \pmstd{0.0466}{0.0101} & \pmstd{\underline{0.0220}}{0.0038} & \pmstd{0.0578}{0.0015} & \pmstd{0.0251}{0.0004} & \pmstd{0.0767}{0.0052} & \pmstd{0.0313}{0.0023} & \pmstd{0.0947}{0.0057} & \pmstd{0.0503}{0.0036} & \pmstd{0.0252}{0.0008} & \pmstd{0.0179}{0.0007} & \pmstd{0.0516}{0.0038} & \pmstd{0.0149}{0.0012} \\
        TiRex & \pmstd{\underline{0.0267}}{0.0016} & \pmstd{0.0232}{0.0009} & \pmstd{\underline{0.0355}}{0.0016} & \pmstd{\underline{0.0154}}{0.0009} & \pmstd{\underline{0.0554}}{0.0057} & \pmstd{\underline{0.0268}}{0.0053} & \pmstd{\underline{0.0690}}{0.0063} & \pmstd{\underline{0.0381}}{0.0039} & \pmstd{\mathbf{0.0125}}{0.0015} & \pmstd{\mathbf{0.0025}}{0.0011} & \pmstd{\underline{0.0167}}{0.0026} & \pmstd{\underline{0.0142}}{0.0017} \\
        Sundial-Base & \pmstd{0.0549}{0.0044} & \pmstd{0.0324}{0.0020} & \pmstd{0.1063}{0.0003} & \pmstd{0.0840}{0.0009} & \pmstd{0.1131}{0.0028} & \pmstd{0.0825}{0.0015} & \pmstd{0.1310}{0.0021} & \pmstd{0.0993}{0.0008} & \pmstd{0.0419}{0.0017} & \pmstd{0.0308}{0.0009} & \pmstd{0.0933}{0.0016} & \pmstd{0.0701}{0.0010} \\
        Aurora & \pmstd{0.0860}{0.0057} & \pmstd{0.0537}{0.0032} & \pmstd{0.1701}{0.0005} & \pmstd{0.1528}{0.0003} & \pmstd{0.1824}{0.0025} & \pmstd{0.1511}{0.0015} & \pmstd{0.1967}{0.0019} & \pmstd{0.1681}{0.0011} & \pmstd{0.0689}{0.0015} & \pmstd{0.0576}{0.0009} & \pmstd{0.1555}{0.0023} & \pmstd{0.1345}{0.0011} \\
        MIRA-Large & \pmstd{0.1101}{0.0035} & \pmstd{0.0949}{0.0024} & \pmstd{0.1782}{0.0005} & \pmstd{0.1627}{0.0006} & \pmstd{0.1926}{0.0018} & \pmstd{0.1710}{0.0015} & \pmstd{0.1999}{0.0013} & \pmstd{0.1766}{0.0005} & \pmstd{0.1542}{0.0017} & \pmstd{0.1473}{0.0016} & \pmstd{0.1894}{0.0011} & \pmstd{0.1735}{0.0009} \\
        \midrule
        \rowcolor[HTML]{E5E5E5}
        SOTER & \pmstd{\mathbf{0.0253}}{0.0007} & \pmstd{\mathbf{0.0205}}{0.0003} & \pmstd{\mathbf{0.0183}}{0.0002} & \pmstd{\mathbf{0.0083}}{0.0001} & \pmstd{\mathbf{0.0303}}{0.0009} & \pmstd{\mathbf{0.0142}}{0.0005} & \pmstd{\mathbf{0.0391}}{0.0005} & \pmstd{\mathbf{0.0182}}{0.0004} & \pmstd{\underline{0.0204}}{0.0003} & \pmstd{\underline{0.0132}}{0.0003} & \pmstd{\mathbf{0.0152}}{0.0004} & \pmstd{\mathbf{0.0091}}{0.0001} \\
        \bottomrule
    \end{tabular}%
    }
\end{table*}

\begin{table*}[t!]
    \caption{Complete continuous-time imputation results at a $50\%$
    missing rate. Reporting conventions follow
    Table~\ref{tab:e-imp25}.}
    \label{tab:e-imp50}
    \centering
    \footnotesize
    \renewcommand{\arraystretch}{1.0}
    \setlength{\tabcolsep}{2.5pt}
    \resizebox{\textwidth}{!}{%
    \begin{tabular}{lcccccccccccc}
        \toprule
        \multirow{2}{*}{Method} & \multicolumn{2}{c}{\shortstack{HeartRate\\(SC)}} & \multicolumn{2}{c}{\shortstack{MIT-BIH\\(MC)}} & \multicolumn{2}{c}{\shortstack{ScientiSST\\MOVE (MC)}} & \multicolumn{2}{c}{\shortstack{IEEE PPG\\(MC)}} & \multicolumn{2}{c}{\shortstack{MMASH\\(MC)}} & \multicolumn{2}{c}{\shortstack{WDD\\(MC)}} \\
        \cmidrule(lr){2-3}\cmidrule(lr){4-5}\cmidrule(lr){6-7}\cmidrule(lr){8-9}\cmidrule(lr){10-11}\cmidrule(lr){12-13}
        & RMSE & MAE & RMSE & MAE & RMSE & MAE & RMSE & MAE & RMSE & MAE & RMSE & MAE \\
        \midrule
        TimeMoE-Large & \pmstd{0.1877}{0.0027} & \pmstd{0.1485}{0.0047} & \pmstd{0.2738}{0.0006} & \pmstd{0.2645}{0.0011} & \pmstd{0.2761}{0.0116} & \pmstd{0.2529}{0.0122} & \pmstd{0.2908}{0.0010} & \pmstd{0.2699}{0.0013} & \pmstd{0.1325}{0.0055} & \pmstd{0.1227}{0.0047} & \pmstd{0.2521}{0.0016} & \pmstd{0.2371}{0.0019} \\
        Moirai2.0-Small & \pmstd{0.1519}{0.0459} & \pmstd{0.1122}{0.0272} & \pmstd{0.2083}{0.0081} & \pmstd{0.1657}{0.0079} & \pmstd{0.1739}{0.0104} & \pmstd{0.1177}{0.0097} & \pmstd{0.2574}{0.0055} & \pmstd{0.2086}{0.0055} & \pmstd{0.0615}{0.0020} & \pmstd{0.0434}{0.0039} & \pmstd{0.0766}{0.0082} & \pmstd{0.0500}{0.0041} \\
        MOMENT-Large & \pmstd{0.0928}{0.0046} & \pmstd{0.0627}{0.0020} & \pmstd{0.2331}{0.0195} & \pmstd{0.2144}{0.0267} & \pmstd{0.1958}{0.0246} & \pmstd{0.1638}{0.0258} & \pmstd{0.2729}{0.0206} & \pmstd{0.2465}{0.0223} & \pmstd{0.0781}{0.0043} & \pmstd{0.0628}{0.0029} & \pmstd{0.2742}{0.0153} & \pmstd{0.2490}{0.0159} \\
        Chronos2-Base & \pmstd{0.1357}{0.0024} & \pmstd{0.0982}{0.0026} & \pmstd{0.3028}{0.0038} & \pmstd{0.2811}{0.0035} & \pmstd{0.3112}{0.0133} & \pmstd{0.2690}{0.0114} & \pmstd{0.3430}{0.0078} & \pmstd{0.3078}{0.0063} & \pmstd{0.1214}{0.0041} & \pmstd{0.1036}{0.0032} & \pmstd{0.2667}{0.0085} & \pmstd{0.2287}{0.0053} \\
        TiRex & \pmstd{\underline{0.0462}}{0.0029} & \pmstd{\underline{0.0240}}{0.0022} & \pmstd{\underline{0.0442}}{0.0020} & \pmstd{\underline{0.0198}}{0.0008} & \pmstd{\underline{0.0800}}{0.0027} & \pmstd{\underline{0.0361}}{0.0030} & \pmstd{\underline{0.0993}}{0.0033} & \pmstd{\underline{0.0560}}{0.0027} & \pmstd{\mathbf{0.0166}}{0.0011} & \pmstd{\mathbf{0.0056}}{0.0010} & \pmstd{\underline{0.0297}}{0.0007} & \pmstd{\underline{0.0202}}{0.0012} \\
        Sundial-Base & \pmstd{0.0997}{0.0185} & \pmstd{0.0761}{0.0106} & \pmstd{0.2737}{0.0007} & \pmstd{0.2594}{0.0001} & \pmstd{0.2806}{0.0057} & \pmstd{0.2471}{0.0062} & \pmstd{0.3007}{0.0035} & \pmstd{0.2751}{0.0024} & \pmstd{0.1078}{0.0032} & \pmstd{0.0969}{0.0035} & \pmstd{0.2412}{0.0041} & \pmstd{0.2191}{0.0032} \\
        Aurora & \pmstd{0.1171}{0.0075} & \pmstd{0.0839}{0.0074} & \pmstd{0.2877}{0.0025} & \pmstd{0.2765}{0.0019} & \pmstd{0.2949}{0.0054} & \pmstd{0.2628}{0.0065} & \pmstd{0.3178}{0.0041} & \pmstd{0.2953}{0.0028} & \pmstd{0.1152}{0.0020} & \pmstd{0.1035}{0.0021} & \pmstd{0.2574}{0.0036} & \pmstd{0.2387}{0.0031} \\
        MIRA-Large & \pmstd{0.1249}{0.0345} & \pmstd{0.1008}{0.0216} & \pmstd{0.2714}{0.0019} & \pmstd{0.2609}{0.0015} & \pmstd{0.2754}{0.0044} & \pmstd{0.2498}{0.0052} & \pmstd{0.2945}{0.0026} & \pmstd{0.2749}{0.0010} & \pmstd{0.1748}{0.0033} & \pmstd{0.1550}{0.0037} & \pmstd{0.2532}{0.0042} & \pmstd{0.2383}{0.0034} \\
        \midrule
        \rowcolor[HTML]{E5E5E5}
        SOTER & \pmstd{\mathbf{0.0280}}{0.0003} & \pmstd{\mathbf{0.0177}}{0.0010} & \pmstd{\mathbf{0.0307}}{0.0009} & \pmstd{\mathbf{0.0177}}{0.0006} & \pmstd{\mathbf{0.0459}}{0.0043} & \pmstd{\mathbf{0.0256}}{0.0019} & \pmstd{\mathbf{0.0520}}{0.0009} & \pmstd{\mathbf{0.0295}}{0.0016} & \pmstd{\underline{0.0262}}{0.0004} & \pmstd{\underline{0.0192}}{0.0003} & \pmstd{\mathbf{0.0270}}{0.0010} & \pmstd{\mathbf{0.0195}}{0.0003} \\
        \bottomrule
    \end{tabular}%
    }
\end{table*}

\begin{table*}[t!]
    \caption{Complete continuous-time imputation results at a $75\%$
    missing rate. Reporting conventions follow
    Table~\ref{tab:e-imp25}.}
    \label{tab:e-imp75}
    \centering
    \footnotesize
    \renewcommand{\arraystretch}{1.0}
    \setlength{\tabcolsep}{2.5pt}
    \resizebox{\textwidth}{!}{%
    \begin{tabular}{lcccccccccccc}
        \toprule
        \multirow{2}{*}{Method} & \multicolumn{2}{c}{\shortstack{HeartRate\\(SC)}} & \multicolumn{2}{c}{\shortstack{MIT-BIH\\(MC)}} & \multicolumn{2}{c}{\shortstack{ScientiSST\\MOVE (MC)}} & \multicolumn{2}{c}{\shortstack{IEEE PPG\\(MC)}} & \multicolumn{2}{c}{\shortstack{MMASH\\(MC)}} & \multicolumn{2}{c}{\shortstack{WDD\\(MC)}} \\
        \cmidrule(lr){2-3}\cmidrule(lr){4-5}\cmidrule(lr){6-7}\cmidrule(lr){8-9}\cmidrule(lr){10-11}\cmidrule(lr){12-13}
        & RMSE & MAE & RMSE & MAE & RMSE & MAE & RMSE & MAE & RMSE & MAE & RMSE & MAE \\
        \midrule
        TimeMoE-Large & \pmstd{0.2114}{0.0063} & \pmstd{0.1679}{0.0056} & \pmstd{0.3786}{0.0020} & \pmstd{0.3721}{0.0019} & \pmstd{0.3787}{0.0076} & \pmstd{0.3492}{0.0088} & \pmstd{0.4027}{0.0045} & \pmstd{0.3855}{0.0038} & \pmstd{0.1649}{0.0047} & \pmstd{0.1498}{0.0064} & \pmstd{0.3388}{0.0014} & \pmstd{0.3206}{0.0020} \\
        Moirai2.0-Small & \pmstd{0.1808}{0.0398} & \pmstd{0.1490}{0.0333} & \pmstd{0.2885}{0.0171} & \pmstd{0.2471}{0.0185} & \pmstd{0.2338}{0.0117} & \pmstd{0.1713}{0.0112} & \pmstd{0.3516}{0.0038} & \pmstd{0.3072}{0.0061} & \pmstd{0.0883}{0.0039} & \pmstd{0.0691}{0.0021} & \pmstd{\underline{0.1378}}{0.0063} & \pmstd{\underline{0.0881}}{0.0087} \\
        MOMENT-Large & \pmstd{0.1328}{0.0179} & \pmstd{0.1000}{0.0113} & \pmstd{0.3510}{0.0302} & \pmstd{0.3430}{0.0311} & \pmstd{0.3269}{0.0168} & \pmstd{0.2945}{0.0157} & \pmstd{0.3911}{0.0108} & \pmstd{0.3707}{0.0119} & \pmstd{0.1088}{0.0045} & \pmstd{0.0883}{0.0044} & \pmstd{0.3594}{0.0027} & \pmstd{0.3383}{0.0024} \\
        Chronos2-Base & \pmstd{0.2047}{0.0651} & \pmstd{0.1617}{0.0425} & \pmstd{0.4850}{0.0014} & \pmstd{0.4809}{0.0014} & \pmstd{0.4816}{0.0101} & \pmstd{0.4491}{0.0123} & \pmstd{0.5100}{0.0050} & \pmstd{0.4959}{0.0046} & \pmstd{0.1895}{0.0050} & \pmstd{0.1800}{0.0076} & \pmstd{0.4280}{0.0050} & \pmstd{0.4075}{0.0049} \\
        TiRex & \pmstd{\underline{0.0714}}{0.0059} & \pmstd{\underline{0.0392}}{0.0032} & \pmstd{\underline{0.1603}}{0.0036} & \pmstd{\underline{0.1122}}{0.0063} & \pmstd{\underline{0.1722}}{0.0083} & \pmstd{\underline{0.1122}}{0.0063} & \pmstd{\underline{0.1897}}{0.0059} & \pmstd{\underline{0.1093}}{0.0044} & \pmstd{\underline{0.0789}}{0.0025} & \pmstd{\underline{0.0509}}{0.0035} & \pmstd{0.1463}{0.0094} & \pmstd{0.1126}{0.0041} \\
        Sundial-Base & \pmstd{0.1076}{0.0065} & \pmstd{0.0974}{0.0029} & \pmstd{0.4105}{0.0101} & \pmstd{0.4052}{0.0110} & \pmstd{0.4083}{0.0082} & \pmstd{0.3823}{0.0059} & \pmstd{0.4499}{0.0047} & \pmstd{0.4335}{0.0062} & \pmstd{0.1955}{0.0037} & \pmstd{0.1700}{0.0014} & \pmstd{0.3601}{0.0115} & \pmstd{0.3503}{0.0136} \\
        Aurora & \pmstd{0.1642}{0.0254} & \pmstd{0.1160}{0.0124} & \pmstd{0.4030}{0.0021} & \pmstd{0.3966}{0.0027} & \pmstd{0.4069}{0.0186} & \pmstd{0.3759}{0.0199} & \pmstd{0.4322}{0.0008} & \pmstd{0.4149}{0.0026} & \pmstd{0.1881}{0.0072} & \pmstd{0.1649}{0.0087} & \pmstd{0.3605}{0.0014} & \pmstd{0.3405}{0.0027} \\
        MIRA-Large & \pmstd{0.1556}{0.0298} & \pmstd{0.1226}{0.0230} & \pmstd{0.3720}{0.0026} & \pmstd{0.3648}{0.0033} & \pmstd{0.3756}{0.0173} & \pmstd{0.3464}{0.0181} & \pmstd{0.3962}{0.0017} & \pmstd{0.3782}{0.0020} & \pmstd{0.1991}{0.0046} & \pmstd{0.1507}{0.0058} & \pmstd{0.3335}{0.0041} & \pmstd{0.3161}{0.0038} \\
        \midrule
        \rowcolor[HTML]{E5E5E5}
        SOTER & \pmstd{\mathbf{0.0546}}{0.0071} & \pmstd{\mathbf{0.0317}}{0.0020} & \pmstd{\mathbf{0.1502}}{0.0047} & \pmstd{\mathbf{0.0860}}{0.0041} & \pmstd{\mathbf{0.1641}}{0.0050} & \pmstd{\mathbf{0.0915}}{0.0033} & \pmstd{\mathbf{0.1684}}{0.0059} & \pmstd{\mathbf{0.0984}}{0.0039} & \pmstd{\mathbf{0.0541}}{0.0029} & \pmstd{\mathbf{0.0408}}{0.0019} & \pmstd{\mathbf{0.1282}}{0.0042} & \pmstd{\mathbf{0.0730}}{0.0025} \\
        \bottomrule
    \end{tabular}%
    }
\end{table*}

\begin{figure}[!tbp]
    \centering
    \includegraphics[width=\columnwidth]{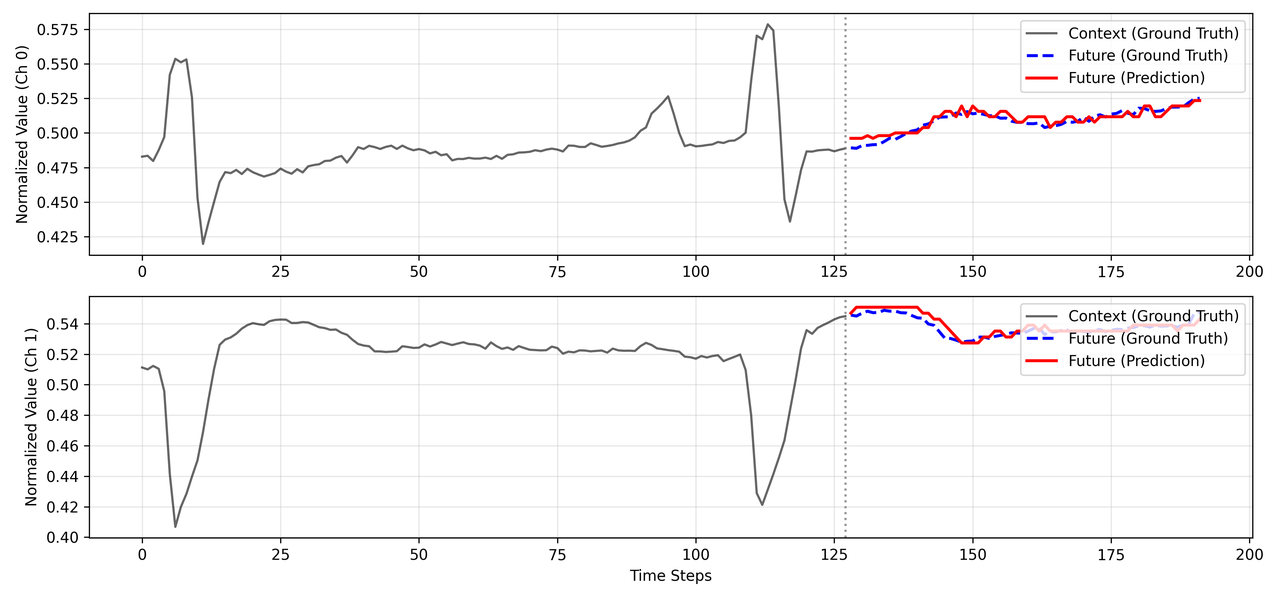}
    \caption{Zero-shot forecasting example on a randomly selected
    two-channel MIT-BIH sample under the $128/64$ input/output setting.
    The gray curves denote the observed context, the blue dashed curves
    denote the ground-truth future, and the red curves denote SOTER's
    predictions. The vertical dotted line marks the forecasting origin.}
    \label{fig:viz-mc}
\end{figure}

\begin{figure}[!tbp]
    \centering
    \includegraphics[width=\columnwidth]{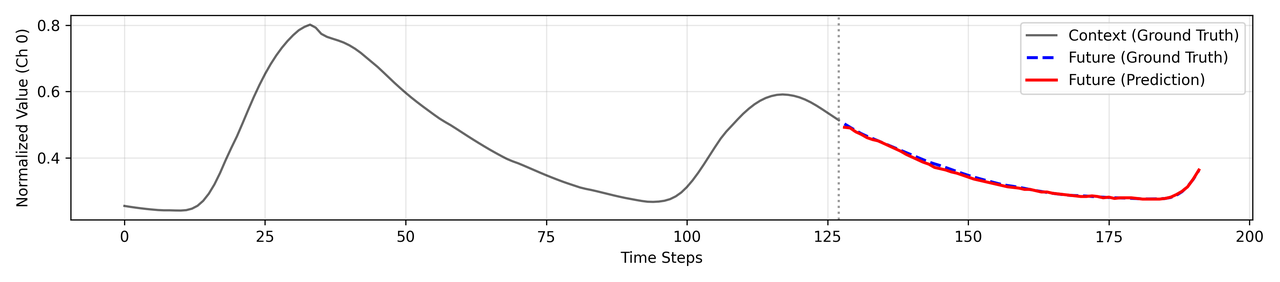}
    \caption{Zero-shot forecasting example on a randomly selected
    single-channel HeartRate sample under the $128/64$ input/output
    setting. The gray curve denotes the observed context, the blue
    dashed curve denotes the ground-truth future, and the red curve
    denotes SOTER's prediction. The vertical dotted line marks the
    forecasting origin.}
    \label{fig:viz-sc}
\end{figure}

\section{Zero-Shot Forecasting Visualizations}
\label{app:forecasting_visualizations}

\begin{table*}[t!]
\centering
\caption{Label-efficient frozen linear probing on ScientiSST MOVE under varying supervision budgets. Results are reported as mean$\pm$std over three runs. Best scores are highlighted by bold style and second-best are underlined.}
\label{tab:label_efficiency}
\setlength{\tabcolsep}{3.0pt}
\renewcommand{\arraystretch}{1.08}
\resizebox{\textwidth}{!}{
\begin{tabular}{lcccccccccccc}
\toprule
\multirow{2}{*}{Methods}
& \multicolumn{6}{c}{Macro-F1}
& \multicolumn{6}{c}{Macro-AUROC} \\
\cmidrule(lr){2-7}\cmidrule(lr){8-13}
& 1\% & 5\% & 10\% & 25\% & 50\% & 100\%
& 1\% & 5\% & 10\% & 25\% & 50\% & 100\% \\
\midrule
TF-C
& \pmstd{31.68}{0.92} & \pmstd{33.65}{0.55} & \pmstd{34.68}{0.51}
& \pmstd{37.26}{0.57} & \pmstd{43.51}{0.60} & \pmstd{47.79}{2.35}
& \pmstd{83.24}{0.43} & \pmstd{\underline{86.33}}{0.69} & \pmstd{87.81}{0.39}
& \pmstd{88.51}{0.52} & \pmstd{90.21}{0.41} & \pmstd{91.89}{2.02} \\

MOMENT-Large
& \pmstd{30.87}{1.56} & \pmstd{34.57}{0.75} & \pmstd{\underline{42.72}}{0.31}
& \pmstd{44.99}{0.67} & \pmstd{47.91}{0.79} & \pmstd{49.60}{0.55}
& \pmstd{81.41}{0.44} & \pmstd{83.88}{0.31} & \pmstd{86.61}{0.36}
& \pmstd{87.78}{0.48} & \pmstd{88.48}{0.70} & \pmstd{88.55}{0.64} \\

NormWear
& \pmstd{\underline{35.48}}{0.81} & \pmstd{\underline{38.91}}{0.49} & \pmstd{40.79}{0.68}
& \pmstd{42.66}{0.38} & \pmstd{42.90}{0.99} & \pmstd{43.03}{0.16}
& \pmstd{\underline{84.19}}{0.55} & \pmstd{85.92}{0.34} & \pmstd{88.36}{0.38}
& \pmstd{\underline{91.21}}{0.40} & \pmstd{\underline{90.64}}{1.14}
& \pmstd{\underline{91.92}}{0.98} \\

MIRA-Large
& \pmstd{32.47}{0.99} & \pmstd{36.80}{0.67} & \pmstd{41.50}{0.51}
& \pmstd{\underline{47.32}}{0.87} & \pmstd{\underline{48.70}}{0.86}
& \pmstd{\underline{50.09}}{1.22}
& \pmstd{83.86}{0.61} & \pmstd{85.38}{0.45} & \pmstd{\underline{88.77}}{0.35}
& \pmstd{91.02}{0.84} & \pmstd{89.69}{0.78} & \pmstd{90.07}{0.81} \\
\midrule
SOTER
& \pmstd{\textbf{40.48}}{0.83} & \pmstd{\textbf{41.69}}{0.45}
& \pmstd{\textbf{44.87}}{0.46} & \pmstd{\textbf{48.01}}{0.88}
& \pmstd{\textbf{50.42}}{0.54} & \pmstd{\textbf{50.90}}{0.36}
& \pmstd{\textbf{89.03}}{0.56} & \pmstd{\textbf{91.38}}{0.45}
& \pmstd{\textbf{93.41}}{0.52} & \pmstd{\textbf{94.89}}{0.43}
& \pmstd{\textbf{94.95}}{0.24} & \pmstd{\textbf{95.62}}{0.12} \\
\bottomrule
\end{tabular}}
\end{table*}

To complement the quantitative results in
Appendix~\ref{app:forecasting_results}, we visualize representative
zero-shot forecasts produced by SOTER under the longest input/output
window, $128/64$. The samples are randomly selected from MIT-BIH and
HeartRate to illustrate forecasting behavior on multi-channel and
single-channel physiological signals, respectively. In both figures,
the gray curve denotes the observed context, the blue dashed curve
denotes the ground-truth future, the red curve denotes SOTER's
prediction, and the vertical dotted line marks the forecasting origin.
Figure~\ref{fig:viz-mc} presents a two-channel MIT-BIH example, whereas
Figure~\ref{fig:viz-sc} presents a single-channel HeartRate example.

\section{Label-Efficient Frozen-Representation Evaluation}
\label{app:label_efficiency}
To examine the label efficiency of the frozen representations, we evaluate linear probing on ScientiSST MOVE using 1\%, 5\%, 10\%, 25\%, 50\%, and 100\% of the labeled training samples, while keeping every pre-trained backbone fixed. As shown in Table~\ref{tab:label_efficiency}, SOTER achieves the best Macro-F1 and Macro-AUROC at every supervision level. The advantage is particularly clear in the low-label regime: with only 1\% of the labeled data, SOTER obtains a Macro-F1 of 40.48 and a Macro-AUROC of 89.03, exceeding the strongest competing results by 5.00 and 4.84 percentage points, respectively. With 25\% of the labels, SOTER already reaches 48.01 Macro-F1 and 94.89 Macro-AUROC, compared with 50.90 and 95.62 under full supervision. These results demonstrate that, on ScientiSST MOVE, SOTER's frozen representations remain effective when downstream labels are limited.

\section{LLM Usage Declaration}
\label{app:llm_declaration}
A large language model (LLM) was used as an auxiliary tool during the
preparation of this work, primarily for code-level assistance and
English-language refinement. The research problem, methodological
design, dataset selection and processing, experimental protocol,
mathematical analysis, interpretation of results, and scientific
conclusions were developed by the human authors. Any LLM-assisted code
or text was reviewed, revised, and verified by the authors, who take
full responsibility for the correctness and integrity of the submitted
work.

\end{document}